\def\year{2021}\relax
\documentclass[letterpaper]{article} % DO NOT CHANGE THIS
\usepackage{aaai21}  % DO NOT CHANGE THIS
\usepackage{times}  % DO NOT CHANGE THIS
\usepackage{helvet} % DO NOT CHANGE THIS
\usepackage{courier}  % DO NOT CHANGE THIS
\usepackage[hyphens]{url}  % DO NOT CHANGE THIS
\usepackage{floatrow}
\usepackage{graphicx} % DO NOT CHANGE THIS
\usepackage{natbib}  % DO NOT CHANGE THIS AND DO NOT ADD ANY OPTIONS TO IT
\usepackage{caption} % DO NOT CHANGE THIS AND DO NOT ADD ANY OPTIONS TO IT
\usepackage{multirow}
\usepackage{color}
\usepackage{colordvi}
\usepackage{amssymb}
\usepackage{amsmath}
\usepackage{enumitem}
\usepackage{parskip}
\usepackage{cleveref}
\usepackage[ruled,vlined]{algorithm2e}
\usepackage{algorithmic}
\usepackage{floatrow}
\newfloatcommand{capbtabbox}{table}[][\FBwidth]
\newcommand{\indep}{\rotatebox[origin=c]{90}{$\models$}}

\title{Meta Learning for Causal Direction}
\author {
        Jean-Fran\c cois Ton,\textsuperscript{\rm 1}
        Dino Sejdinovic, \textsuperscript{\rm 1}
        Kenji Fukumizu \textsuperscript{\rm 2} \\
}
\affiliations {
    \textsuperscript{\rm 1} University of Oxford \\
    \textsuperscript{\rm 2} The Institute of Statistical Mathematics \\
    ton@stats.ox.ac.uk, dino.sejdinovic@stats.ox.ac.uk, fukumizu@ism.ac.jp
}
\begin{document}

\maketitle

\begin{abstract}
The inaccessibility of controlled randomized trials due to inherent constraints in many fields of science has been a fundamental issue in causal inference. In this paper, we focus on distinguishing the cause from effect in the bivariate setting under limited observational data. Based on recent developments in meta learning as well as in causal inference, we introduce a novel generative model that allows distinguishing cause and effect in the small data setting. Using a learnt task variable that contains distributional information of each dataset (task), we propose an end-to-end algorithm that makes use of similar training datasets at test time. We demonstrate our method on various synthetic as well as real-world data and show that it is able to maintain high accuracy in detecting directions across varying dataset sizes.
\end{abstract}

\section{Introduction}
Discovering causal links between variables has been a long standing problem in many areas of science. Ideally all experiments are in a randomized controlled environment where each variable can be accounted for separately. However in most cases, this is impossible due to physical, financial or ethical reasons. The problem of determining causal direction becomes even more apparent, when trying to understand the true generating process of data. Although modern machine learning models are able to achieve impressive performances in capturing complex nonlinear relationships among variables, many of them do not take into account causal structure, which might lead to generalization errors when faced with different data than seen during training.
% when moving from an observational to an interventional regime.

Hence, in recent years, researchers have focused on inferring causal relations from observational data and have developed many different algorithms. One of the major approaches is constraint-based methods, which analyze the conditional independence among variables to determine the causal graph up to a Markov equivalence class under certain assumptions \cite{neuberg2003causality}; in addition to early example of the PC algorithm \cite{spirtes2000causation}, there are also nonlinear methods for capturing independence \cite{sun2007kernelCL,sun2007distinguishing, zhang2012kernel}. 

Another category is score-based methods which use search algorithms to find the best causal graph with respect to such a score as BIC \cite{chickering2002optimal}. These methods are, however, often unable to determine the correct structure and can be computationally very expensive.  There are also hybrid methods which mitigate such difficulty \cite{tsamardinos2006max}.

A new line of research has taken specific interest in the bivariate case, i.e., the cause-effect inference, where one decides between causal hypotheses ``$X\to Y$'' and ``$Y\to X$'' \cite{hoyer2009nonlinear, goudet2017causal, mitrovic2018causal, wu2020causal}. In this setting, methods that exploit the inherent asymmetries between cause and effect are the most prominent. The data is analysed under the \textit{Functional Causal Model} (FCM, \cite{pearl2009causality}) formalism, following respective model assumptions. Due to the intrinsic asymmetry of the problem, several statistics have been proposed to infer the direction of the cause-effect pairs \cite{shimizu2006linear, hoyer2009nonlinear, mooij2016distinguishing, mitrovic2018causal}.  
%We revisit the problem of cause-effect inference in this paper.  

Most relevant prior work to this paper is the framework of Causal Generative Neural Networks (CGNN, \cite{goudet2017causal}).  When applied to cause-effect inference, CGNN learns a generative model using a neural network for each direction and compares their fitting to determine the causal directionality. 
Many advanced methods including CGNN, however, assume an access to a large dataset to make use of strong learning models such as neural networks. This in turn may lead significantly degraded performances in small data settings encountered in practical problems \cite{look2018learning}. We demonstrate this phenomenon in our experimental section. In addition, training a model for every dataset may cause significantly slow inference time by the computational burden of neural networks. 
%has to be trained i.e. training a neural network, for every new dataset, which very slow inference time.

In this paper, we revisit the problem of cause-effect inference from the viewpoint of empirical learning. Specifically, we are interested in learning from many examples of cause-effect datasets together with their true causal directions. 
Our purpose is to develop a method for using this empirical knowledge on causal directions effectively, when making cause-effect inference on a new unseen and purely observational dataset.

Learning-based cause-effect inference has been already explored in the literature. For instance, Randomized Causation Coefficient (RCC) \cite{lopez2015randomized} and Neural Causation Coefficient (NCC) \cite{lopez2017discovering} make a learning-based binary classifier for causal directions.  RCC and NCC,  however, require to synthesize vast amounts of problem specific data pairs up $10000$ \cite{lopez2017discovering}.
Other examples, such as NonSENS \cite{monti2019causal} and Causal Mosaic \cite{wu2020causal}, aim to recover a FCM using nonlinear independent component analysis. An important assumption for these methods is availability of multiple datasets sharing the same causal mechanism and the same exponential family of latent variables.  We then need to assume or select datasets to satisfy this.  

%When using standard causal inference algorithm, we usually assume access to one large dataset. However, due to the rise of meta learning in the past years, researcher have been more and more interested in the case where one is only given a small dataset for each problem. In addition, most methods that make little to no assumptions on the data are very data dependent, in a sense that the performance significantly degrades as the number of data available decreases. Example of which is  the Causal Generative Neural Network \cite{goudet2017causal}, which aims to learn generative models for causal from inference. We show in our experiments that even though very flexible will not perform well with less data.

Different from these works, aiming at alleviating the problem of small data, we consider methods of meta learning by introducing a dataset-feature extractor. The feature represents the distributional information of each dataset, aiming to encode similar causal mechanisms of datasets into similar features.  For this purpose, we employ two approaches: the formalism of kernel mean embeddings \cite{muandet2017kernel} and DeepSets \cite{zaheer2017deep}. 

We propose a neural network-based generative model that trains jointly on all the training datasets.  The model has an encoder-decoder architecture, which has been employed  successfully by meta learning frameworks \cite{garnelo2018neural,garnelo2018conditional}; the encoder gives the dataset-features, and the decoder realizes a generative model or FCM, which is accompanied with the Feature-wise Linear Modulation layers (FiLM  \cite{perez2018film}) to adapt the generator to the dataset at hand.  With this meta-learning architecture, the proposed method is able to determine the cause-effect direction efficiently for new unseen, possibly small, and purely observational datasets.  

%Furthermore we make use of the kernel methods literature such as kernel mean embeddings to capture the asymmetry in the dataset. Several works  have used kernel methods in causal inference \cite{mitrovic2018causal, lopez2015randomized}, however, neither of these methods use them in a generative model, which we show in this paper can greatly benefit from these embeddings.

The contributions of this work can be summarized as follows:
\begin{itemize}
\item We introduce a new meta learning algorithm that can leverage similar datasets for unseen causal pairs in causal direction discovery.
\item We exploit structural asymmetries with an adaptive generative model, thus avoiding the need to retrain at test time.
\item We propose an end-to-end algorithm that assumes no \textit{a priori} assumptions on the causal mechanism between cause and effect.
\item High performance on small dataset sizes can be achieved by virtue of meta learning.
\end{itemize}

\section{Meta Learning for Detecting Causal Direction}
%\section{Background}
%We firstly give a brief summary on techniques and frameworks we will be using through out the paper. In particular we give a brief introduction to kernel mean embeddings \cite{song2013kernel} as they constitute an important component in our methodology.
We first give a brief summary of FCM and explain the proposed method including its building blocks. 

\subsection{Functional Causal Model}
Functional Causal Models (FCM) have been widely used when conducting causal inference. Formally, a FCM on a random vector $X=(X_1, \dots, X_d)$ is a triplet $C=(\mathcal{G}, f, \mathcal{E})$, where $(\mathcal{G}$ is the causal graph and $f, \mathcal{E}$ such that:
\begin{align}
    X_i =  f_i(X_{Pa(i;\mathcal{G})}, Z_i), \quad Z_i \sim \mathcal{E}, \text{  for } i=1, \dots, d,
\end{align}
where $X_i$ are the observed variables, $Z_i$ are the independent hidden variables, $Pa(i;\mathcal{G})$ being the parents of $X_i$ and $f_i$ being the mechanism linking the cause and the effect. Under this formulation, there is clear asymmetry between cause and effect, given that cause is used to infer the effect, and hence numerous work has been done exploiting this fact. 

For inferring causal direction $X\to Y$ for bivariate $(X,Y)$, we can consider only the FCM $Y = f(X,Z)$, where $Z$ and $X$ are independent.  Among other inference methods, CGNN \cite{goudet2017causal} uses neural networks to train the mechanism $f$ for a dataset. More precisely, given dataset $\mathcal{D}=\{(x_j,y_j)\}_{j=1}^m$, we generate $Z_j$ by the standard normal distribution $N(0,1)$, and train neural network $f$ so that the distribution of $\{(X_j,Y_j)\}_j$ be close to that of $\{(X_j, f(X_j,Z_j)\}_j$.  The difference of the distributions is measured by the Maximum Mean Discrepancy (MMD, \cite{gretton2012kernel}).  CGNN learns two models $\hat{Y}=f_y(X,Z)$ and $\hat{X}=f_x(Y,Z)$, and chooses a better fit to determine the direction. 

Unlike CGNN, which trains networks $f_y$ and $f_x$ for each dataset, our method considers a single neural network working for all the datasets in a cause-effect database $\{\mathcal{D}_i\}_{i=1}^N$ where $\mathcal{D}_i = \{(X^i_j,Y^i_j)\}_{j=1}^{m_i}$ is a dataset in the database. We assume that the causal direction is known for all $\mathcal{D}_i$ during training. More specifically, given $X^i\to Y^i$ is the true causal direction, we wish to create a single suitable model $F(X,Z)$ based on neural networks so that the distribution of $\mathcal{D}_i$ is approximately the same as that of $\{(X^i_j,\hat{Y}^i_j)\}$ for {\em any} $i$, where $\hat{Y}^i_j = F(X^i_j,Z^i_j)$. 

This approach involves obvious difficulty, since a wide-variety of cause-effect relations must be learnt by a single network. Na\"ively training a single model jointly over all the different dataset does not yield desired performance as we demonstrate in our ablation study. In order to achieve successful training, we introduce two novel and crucial components: 
\begin{enumerate}
    \item  {\bf Dataset-feature:}  This feature $C$  represents the causal mechanism of each dataset as the  distributional information on the dataset $\mathcal{D}$. The feature will be used to adapt our single network efficiently at test time.
    \item {\bf FiLM layers:} To adapt the base neural network to each dataset using the dataset-features $C$, the FiLM layers enable us to adapt the weights of our network to a given new dataset $\mathcal{D}$ quickly at test time.
\end{enumerate}
Together, with these two additional apparatus, we are able to train our model across datasets and therefore harness information from all the datasets together, instead of treating them independently as it is done for example in CGNN. 

In the next section we will briefly give an high level overview of meta learning and then afterwards describe the ways we are able to capture distributional information of dataset $\mathcal{D}_i$, by leveraging, the well studied area of conditional mean embeddings (CME) \cite{song2013kernel} as well as DeepSets \cite{zaheer2017deep}.

%In this paper we also use the FCM formulation, however, instead of restricting the function class of $f$ \cite{shimizu2006linear,hoyer2009nonlinear,zhang2012identifiability}, we will use Neural Networks in a generative model setting just like in \cite{goudet2017causal}. In particular, we aim to meta-learn the $f$, so that given a new dataset, we are able to determine the mechanism without having to retrain our model and hence also determine the causal direction.

\subsection{Meta Learning}
Meta learning is an ever growing area in machine learning as it allows a model to extract information from similar problems/datasets and use this prior information on new unseen datasets. This is achieved, by sharing statistical strength across several causal inference problems. Standard methods usually require a lot of prior information to be useful in small data setting, i.e. knowing it is a linear model etc. Meta learning however learns this prior information through the meta learning training phase, where during training the model is presented with several small datasets, that allows the model to perform well on new unseen dataset.

Meta learning and multi-task learning differ in the following way:
\begin{itemize}
    \item "\textit{The meta-learning problem: Given data/experience on previous tasks, learn a \textbf{new task} more quickly and/or more proficiently}" (Finn, 2019)
    \item "\textit{The multi-task learning problem: Learn all of the tasks more quickly or more proficiently than learning them independently}"(Finn, 2019)
\end{itemize}
In our setting, we are interested in learning the causal direction on new unseen datasets, given a set of datasets where we know the causal direction. This allows us to make efficient use of the information across all the datasets, which is contrary to most cause-effect methods that treat datasets independently and have to be retrained for each new dataset.

\subsection{Dataset features via DeepSets}
DeepSets \cite{zaheer2017deep} have been used as task embedding $C_i$ in previous meta learning literature \cite{garnelo2018conditional, garnelo2018neural,xu2019metafun}.  Using a neural network $\phi_{x,y}$, the task embedding is defined by 
\begin{align}
\label{deepset}
    C_i = \frac{1}{m_i}\sum_{j=1}^m\phi_{x, y}([x_j, y_j]).
\end{align}
DeepSets is a simple flexible approach to encoding sets into vectors, which is also permutation invariant. The latter is important as we do not want the embeddings to change solely based on the order of the elements in the dataset. \citet{zaheer2017deep} show that DeepSets is a universal approximating for any set function. Hence this aggregation method allows us to have a good representation of the dataset.

However, given that we use a concatenation in DeepSets we do not encode the conditional distribution information but rather the joint. Therefore, in this paper, we in addition also consider conditional mean embeddings as dataset-features.

\subsection{Dataset features via Conditional Mean Embeddings (CME)}
\label{sec:CME}
% Next, we will discuss an elegant way to represent full conditional distributions through kernel methods. These representations will be one of the keys to designing dataset specific embeddings that contain useful distributional information.
Kernel mean embeddings of distributions provide a powerful framework for representing probability distributions \cite{song2013kernel, muandet2017kernel}. Given sets $\mathcal{X}$ and $\mathcal{Y}$, with a distribution $P$ over the random variables $(X, Y)$ taking values in $\mathcal{X} \times \mathcal{Y}$, the conditional mean embedding (CME) of the conditional density $p(y|x)$, is defined as:
\begin{align}
    \mu_{Y|X=x} := \mathbb{E}_{Y|X=x}[\phi_y(Y)] = \int_{\mathcal{Y}} \phi_y(y) p(y|x)dy.
    \label{defcme}
\end{align}
where $\phi_y$ is the feature map associated to the reproducing kernel Hilbert space (RKHS) of $Y$, $\mathcal{H}_Y$.
Intuitively, the equation above allows us to represent a probability distribution $p(y|x)$ in a function space such as a RKHS, by taking the expectation under $p(y|x)$ of the features $\phi_y(y) \in \mathcal{H}_Y$.
Hence, for each value of the conditioning variable $x$, we obtain $\mu_{Y|X=x}\in\mathcal H_Y$. 

Following \cite{song2013kernel}, the CME can be associated with the operator $\mathcal{C}_{Y|X}:\mathcal H_X \to \mathcal H_Y$,  known as the conditional mean embedding operator (CMEO), which satisfies
\begin{align}
\mu_{Y|X=x} = \mathcal{C}_{Y|X} \phi_x(x)
\label{cme}
\end{align}
where $\mathcal{C}_{Y|X}:= \mathcal{C}_{YX} \mathcal{C}_{XX}^{-1}$
with $\mathcal{C}_{YX}:=\mathbb{E}_{Y,X}[\phi_y(Y) \otimes \phi_x(X)]$ and $\mathcal{C}_{XX}:=\mathbb{E}_{X,X}[\phi_x(X) \otimes \phi_x(X)]$.  The operator inverse should be understood as a regularized inverse, unless the rigorous inverse exists. 

As a result, the finite sample estimator of $\mathcal{C}_{Y|X}$ based on the dataset $\{(x_j, y_j) \}_{j=1}^n$ can be written as
\begin{align}
\widehat{\mathcal{C}}_{Y|X} = \Phi_y (K + \lambda I)^{-1} \Phi_x^T
\label{CMEO}
\end{align}
where $\Phi_y := (\phi_y(y_1), \dots, \phi_y(y_n))$ and $\Phi_x := (\phi_x(x_1), \dots, \phi_x(x_n))$ are the feature matrices, $K := \Phi_x^T\Phi_x$ is the kernel matrix with entries $K_{i,j}=k_x(x_i,x_j):=\langle \phi_x(x_i),\phi_x(x_j)\rangle$, and $\lambda>0$ is a regularization parameter.

Hence $\widehat{\mu}_{Y|X=x} = \widehat{\mathcal{C}}_{Y|X} \phi_x(x)$ simplifies to a weighted sum of the feature maps of the observed points $y_i$:
\begin{align}
\widehat{\mu}_{Y|X=x} &= \sum_{i=1}^n \beta_i(x) \phi_y(y_i) = \Phi_y \beta(x), \\
\beta(x) &= (\beta_1(x), \dots , \beta_n(x))^T = (K + \lambda I)^{-1}K_{:x},
\end{align}
where $K_{:x} = (k_x(x, x_1), \dots, k_x(x, x_n))^T$.

In fact, when using finite-dimensional feature maps, the conditional mean embedding operator is simply a solution to a vector-valued ridge regression problem (regressing $\phi_y(y)$ to $\phi_x(x)$), which allows computation scaling linearly in the number of observations $n$. 

The Woodbury matrix identity allows us to have computations of either order $\mathcal{O}(n^3)$ or $\mathcal{O}(d^3)+\mathcal{O}(d^2n)$, where $d$ is the dimension of the feature map $\phi_x$. In our case, given that we are in the meta learning setting, the dataset size $n$ is usually rather small and hence the CME can be efficiently computed.

The CME is a canonical way for capturing conditional densities and thus the mechanism in a functional causal model. Therefore the CME also encodes the causal direction. We give further motivations on why we use CMEO as dataset-features in the next few sections.

\subsection{A Feature-wise Linear Modulation (FiLM)}
% we could make a network $F(X,C,Z)$ which is to be applicable to all the datasets.  However, the data-set feature tends to be high-dimensional, and processing all the variety of datasets would be extremely difficult. Alternatively,

% To use the dataset-feature in FCM, we propose an architecture to adapt the network $f(X,Z)$ with dataset-feature by using 
% Feature-wise Linear Modulation (FiLM) layers \cite{perez2018film}.

We propose an architecture that is able to adapt the network $f(X,Z)$, i.e.~a FCM, with the dataset-feature by using the  Feature-wise Linear Modulation (FiLM) \cite{perez2018film}.

The FiLM layers are known to allow network adaptation to new environments quickly without adding further model parameters. They have been shown to work effectively in various tasks of  computer vision  \cite{perez2018film} and regression \cite{requeima2019fast}.  In essence, the FiLM layers work as follows: given a conditioning variable $C$ (this may be the label for image classification) and $l_a$ being the $a^{th}$ layer of a network, the FiLM layer $FL_a$, constructed by a neural network, adapts $l_a$ to $l_a^{FL}$ by
\begin{align}\label{filmlayer}
    (\beta_a, \gamma_a) = FL_a(C),\qquad     l_a^{FL} = \beta_a + \gamma_a \circ l_a,
\end{align}
where $\circ$ is the element-wise multiplication. 
Intuitively, the FiLM layer learns shift and scale parameters, conditioned on $C$, for any given layer $l_a$.  We shall use the CME or DeepSets for $C$.

% In this paper we are specifically interested in the latter, though the first two could also be applied to causal discovery as well. However, we leave this as future work. In essence, Encoder-Decoder based meta learning methods, first of all encode each dataset/task into a task feature using a permutation invariant Encoder. Example of which are Deep Sets \cite{zaheer2017deep}, which first applies a feature map on each data point before taking the mean/max of these features. This task feature is then supposed to represent the dataset in a vector form. Once this task feature is learnt, it can then be used to adapt the decoder for each task separately. At test time, the decoder is then able to perform well on unseen tasks. [ADD an illustration of this]

% \begin{itemize}
%     \item \textit{\textbf{CGNN} in detail done ish}
%     \item \textit{\textbf{meta learning with task embedding features (NP)}} DONE
%     \item \textit{\textbf{CME} done ish}
%     \item \textit{\textbf{MMD} done}
%     \item \textit{\textbf{Film layers}} Done
% \end{itemize}

\subsection{Proposed method}
\label{section:method}
\begin{figure*}[t]
    \centering
    \includegraphics[width=0.55\textwidth]{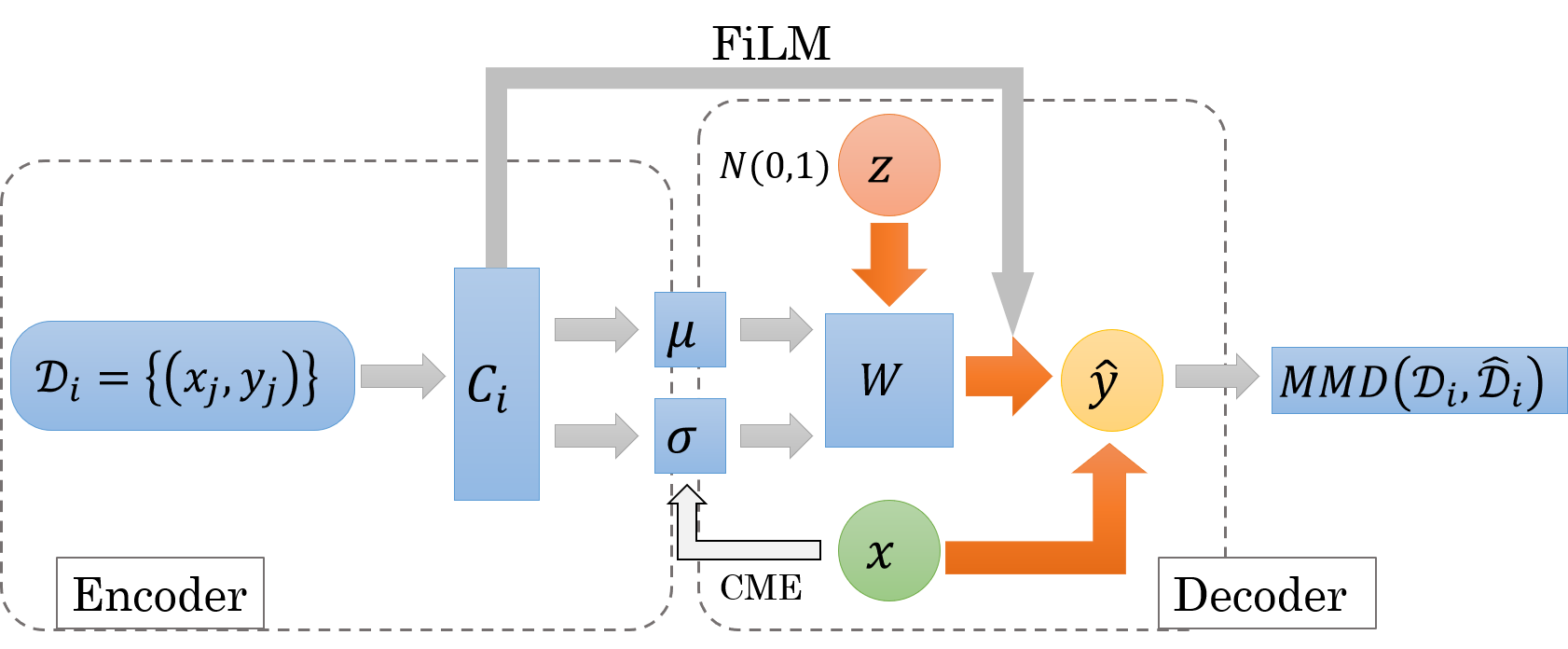}
    \caption{Proposed meta-CGNN Algorithm for only one dataset $\mathcal{D}_i$ in the mini-batch}
    \label{fig:algo1}
\end{figure*}

We propose a new meta learning algorithm, {\em meta-CGNN}, which works for cause-effect inference given cause-effect training database $\mathbb{D}=\{\mathcal{D}_i\}_{i=1}^N$ with known directionality, where $\mathcal{D}_i = \{(x^i_j, y^i_j)\}_{j=1}^{m_i}$.  Without loss of generality, we assume $X^i \rightarrow Y^i$ for any dataset $i$.  The FCM is trained on this database $\mathbb{D}$, and is used to infer the causal direction for an unseen dataset $\mathcal{D}_{test}$.

\textbf{Overview:} We use the popular encoder-decoder based architecture for meta learning, which is similar to the Neural Process \cite{garnelo2018neural}. The encoder first maps the dataset $\mathcal{D}_i$ into a dataset-feature $C_i$, which is given as an input to two further neural networks, (1) FiLM network and (2) amortization network. The FiLM network, operates as described in the above, by producing shift and scale parameters that allow us to adapt the decoder $D_{FL}$ accordingly. The amortization network outputs  $(\mu(C),\sigma(C))$, with which we use to modulate the latent random variable $Z^i_j\sim N(0,1)$ to $W^i_j := \mu(C_i) + \sigma(C_i)Z^i_j$. 

The decoder network, $D_{FL}$, then maps $(X,W)$ to $\hat{Y}$ so that the distribution of $\{(X^i_j,\hat{Y}^i_j)\}_j$ is close to $\{(X^i_j,Y^i_j)\}_j$. By using an encoder network, which trains across datasets jointly, we are able to share distributional information between datasets (tasks) and apply it to a new unseen task. See Figure \ref{fig:algo1} and Algorithm 1 for a detailed breakdown of \textit{meta-CGNN}.
%
%In particular, we are interested in learning a flexible Functional Causal Model (FCM) of the data without imposing any functional constraints. Much of previous work \cite{shimizu2006linear,hoyer2009nonlinear,zhang2012identifiability} assume a certain structure that restricts the functional relationship between cause and effect. The benefit of these constraints is that theoretical guarantees to those methods can be established. However, restricting these model assumptions can also be detrimental, especially if these conditions are not met by the data, i.e. for example additive noise \cite{hoyer2009nonlinear} etc, see our experimental section for more evidence. 
%
%Similarly to \cite{goudet2017causal} we use neural network for the generative model to capture the functional relationship between the cause and effect. Namely, the neural network learns $f$ in the FCM
%This method does not impose any restrictions on the FCM, which makes it applicable to any type of data. The main idea of using generative models in causal inference is to learn the function $f$:

The overall functional causal model in the proposed meta-CGNN is thus
\begin{align}
\label{eq:FCM_task}
    \hat{y}_j = F\left((x_j, z_j); C\right) \text{  , where } z_j \sim N(0, 1), \quad z_j \indep\, x_j.
\end{align}
%Intuitively, if the true causal direction is $X \rightarrow Y$ then learning $y = f(x, z)$ should be easier than $x = g(y, z)$, given that the nature of cause and effect. Hence by exploiting this asymmetry, we are able to detect the causal direction by determining whether $f$ or $g$ fits the data better.
%
%
%However, in general learning $f$ requires a lot of data as generative models are  very data hungry. Therefore, in order to retain a good approximation to the true FCM, we will use a meta learning framework.

% In particular, we aim to tackle firstly, the problem when given very little data  at test time and secondly asked to perform on new problems that are structurally similar but not the same as the ones seen during training. Standard causal inference algorithms usually require vast amounts of data to work well and suffer significant performance loss/ do not work in the small data set setting. In addition, these models have to be also retrained for every new dataset.

\textbf{Encoder:} For representing dataset-specific distributional information or mechanism for each task, we consider both the CME and DeepSets approach.

As argued by \citet{mitrovic2018causal}, CMEs hold critical information for causal-effect inference by representing the mechanism in the FCM. The claim is that the Kolmogorov complexity \cite{grunwald2008algorithmic} of the mechanism tends to be larger in the anti-causal direction than it is in the causal direction. %They use a variety of techniques to exploit this fact. 
Hence, we expect CME to capture relevant distributional information, and inform adapting our generative model to the task at hand.

More concretely, we use conditional mean embeddings as follows. We first compute the CMEO $\mathcal{C}_{Y|X}$ from $\mathcal{D}_i$ as described in (Eq \ref{CMEO}), and use it to obtain the CME for each datapoint using:
% We then create $W_j^i=\mu(C_{i, j})+\sigma(C_{i, j}) Z^i_j$, where $C_{i, j}$ is defined for data pair $(x_j, y_j)$ by 
\begin{align}
\label{cmeij}
    C_{i, j} = \mathcal{C}_{Y|X} \phi_x(x_j).
\end{align}
where, $\phi_x$ is the feature vector such that $k(x_i, x_k)=\langle \phi_x(x_j),\phi_x(x_k) \rangle$, $k$ being the RBF kernel. To obtain a finite dimensional representation efficiently, we will use Random Fourier Features \cite{rahimi2008random} to approximate the CME. Throughout the paper we will use $100$ features as we work on problems with at most $1500$ datapoints. According to \cite{li2018towards}, we need approximately $\sqrt{N}$, where $N$ is the number of datapoints (see Algorithm 1). For DeepSets the dataset feature $C_i$ is simply defined in (Eq \ref{deepset}).

%We decide to use CME as task features in addition to DeepSets, as noted by  \cite{mitrovic2018causal}, CME hold very critical information for causal direction detection. In particular \cite{mitrovic2018causal} show that the Kolmogorov complexity in the anti-causal direction is larger than in the true direction. They use a variety of techniques to exploit this fact. Hence, CME are a sensible way of capturing distributional information, which can thus help us to inform our generative model i.e. Decoder.

\textbf{Decoder:} For the generative part of our model, the dataset-feature $C_i$ or $C_{i, j}$ gives modulation through the FiLM and amortization network. Both FiLM and amortization networks take as input $C_i$ (DeepSets) or $C_{i,j}$ (CME). The FiLM layer is able to adapt the weights of the decoder network depending on the distributional feature of a dataset. This is crucial for a single network to learn FCMs of all the datasets. Na\"ively training a single network over multiple datasets resulted in poor performance (see ablation study). The amortization network works on $z\sim N(0,1)$ similarly to FiLM. It can be interpreted as the adaption of latent Gaussian distribution; $p(w|C_i)$ with $W^i=\mu(C_i)+\sigma(C_i)Z$ regarded as a new latent variable for the dataset $\mathcal{D}_i$. 

%Contrary the the CGNN which has $z \sim N(0, 1)$ for every task, we now have a different source distribution for each separate task. Note however, that this formulation still corresponds to a FCM given that the $\mu_z$ and $\sigma_z$ of the $Z$ are only affine transformation and can be absorbed into $D_{FL}$, leaving us with $Z$ being independent of the cause.
%\begin{align}
%    \widehat{y}_j^k = D_{FL}([x_j, z]) \text{   ,where } z \sim N(0, 1)
%\end{align}
Together with the FiLM layer we construct a decoder which generates data from the conditional distribution, by firstly sampling from $p(w|\mathcal{D}_i)$ and concatenate $w$ with $x$ before pushing it through the decoder $D_{FL}$. This novel architecture, allows us to model and more importantly, \textbf{sample} from the conditional distribution of unseen task quickly and efficiently. In the next section, we will describe how we will make use of the samples to train our networks.

\begin{align}
\label{eq:FCM_task}
    \hat{y}_j = D_{FL}\left((x_j, w_j); C\right) \text{  , where } w_j \sim N(\mu(C), \sigma(C))
\end{align}

\textbf{Training:} The objective function of training is similar to \cite{goudet2017causal}; by sampling $\{\hat{y}^i_j\}_{j=1}^{m_i}$ from \eqref{eq:FCM_task} and estimating the Maximum Mean Discrepancy (MMD) \cite{gretton2012kernel} between the sampled data $\widehat{\mathcal{D}_i} = \{(x^i_j,\hat{y}^i_j)\}_{j=1}^{m_i}$ and the original data $\mathcal{D}_i = \{(x^i_j,y^i_j)\}_{j=1}^m$. MMD is a popular metric to measure the distance between two distributions. It has been widely used in two-sample tests \cite{gretton2012kernel} as well as in applications of Generative Adversarial Network (GAN) \cite{li2017mmd} etc.

More formally, the MMD estimator between two datasets $\mathcal{U}=\{u_i\}_{i=1}^m$ and $\mathcal{V}=\{v_i\}_{i=1}^n$ is, given by
%
% Given $m-$samples $\{x_i\}_{i=1}^m$ of a distribution $\mathcal{P}$ and  $n-$samples $\{y_i\}_{i=1}^n$ of a distribution $\mathcal{Q}$, the population MMD is defined as:
\begin{align*}
    \widehat{\text{MMD}}^2(\mathcal{U}, \mathcal{V}) &= 
    \frac{1}{m(m-1)} \sum_{i=1}^m\sum_{j\neq i}^m k(u_i, u_j)\\ &+ \frac{1}{n(n-1)} \sum_{i=1}^n\sum_{j\neq i}^n k(v_i, v_j)\\&- \frac{2}{mn} \sum_{i=1}^m\sum_{j=1}^n k(u_i, v_j).
\end{align*}
We use Gaussian kernel, which is characteristic \cite{sriperumbudur2010universality}.
%This is a good non-parametric distance metric as the MMD is equal to zero if and only if $\mathcal{P}=\mathcal{Q}$ as $n, m$ go to infinity. 
With $k$ being the Gaussian kernel, the above expression is differentiable and thus can be optimized as already demonstrated in various works such as \cite{goudet2017causal, li2017mmd}. A drawback of using MMD as a loss function however is that it scales quadratically in the sample size.  We can again use Random Fourier Features \cite{rahimi2008random, lopez2015randomized}, which give us linear-time estimators of MMD.
%by approximating the Kernel matrix using random projections. This allows one to cut the computation cost to $\mathcal{O}(m^2)$ where $m$ is the number of RFF such that $m << N$. \cite{li2018towards} show that $m$ only needs to be $\mathcal{O}(\sqrt{N})$.

Using stochastic mini-batches of $q$ datasets $\{\mathcal{D}_i\}_{i=1}^q$, the objective function to minimize is
\begin{align}\label{loss}
    LOSS = \sum_{i=1}^q \widehat{\text{MMD}}^2(\widehat{\mathcal{D}_i}, \mathcal{D}_i)
\end{align}
This joint training, similar to that used in other encoder-decoder architectures for meta learning \cite{garnelo2018conditional, garnelo2018neural, xu2019metafun}, allows us to utilize the information of all the available training datasets in a single generative model. This is in stark contrast to CGNN \cite{goudet2017causal}, which trains a separate generative model for each dataset. As noted in \cite{goudet2017causal} training these neural networks separately can be very costly and one of the major drawbacks of the CGNN. Hence \textit{meta-CGNN} aims to alleviate this constraint by training over datasets jointly. This considerably speeds up the model inference time as \textit{meta-CGNN} does not need to retrain the network for a new task.

%  and has been shown to achieve state-of-the-art performances on numerous meta learning tasks

\textbf{Inference of Causal Direction:} After training, when we wish to infer the causal direction for a new dataset $\mathcal{D}_{test}=\{(x_j,y_j)\}_{j=1}^m$, we feed both of  $\mathcal{D}^{xy} = \{(x_j,y_j)\}_{j=1}^m$ and $\mathcal{D}^{yx} = \{(y_j,x_j)\}_{j=1}^m$ into the trained model and estimate the MMDs between the generated samples and the true ones, i.e., $\mathcal{M}_{xy}:=\widehat{\text{MMD}}(\mathcal{D}^{xy},\hat{\mathcal{D}}^{x\hat{y}})$ and $\mathcal{M}_{yx}:=\widehat{\text{MMD}}(\mathcal{D}^{yx},\hat{\mathcal{D}}^{y\hat{x}})$. If $\mathcal{M}_{xy} < \mathcal{M}_{yx}$, we deduce that $X\to Y$ as it agrees better with a postulated FCM than $Y\to X$, and similarly $Y\to X$ if $\mathcal{M}_{xy} > \mathcal{M}_{yx}$.
Intuitively, this means that we choose the direction of whose samples that match the ground truth best.

% \begin{itemize}
%     % \item how do we do task embedding? DONE
%     % \item how do we modify the CGNN to make it task depdendent? DONE
%     % \item added FILM layer (need to do ablation study on this) DONE
%     % \item added different normal sources with different variances and means compared to std gaussian in CGNN DONE
%     % \item training jointly and using previous knowledge of training data to compute the direction DONE
%     %\item add figure of the algorithm
%     \item add the theory that was given by CGNN as it directly applies to our method as well
% \end{itemize}

\section{Related work}
There have been already some works that exploit the asymmetry by taking a closer look at decomposing the joint distribution $P(X, Y)$ into either $P(Y|X)P(X)$ or $P(X|Y)P(Y)$. 
Relevant to this work is the approach using the asymmetry in terms of functional causal models (FCMs).  Some of the previous methods make strong assumptions on the model; LiNGAM \cite{shimizu2006linear} considers linear non-Gaussian model for finding causal structure.  For nonlinear relations, \citet{hoyer2009nonlinear} discusses nonlinear additive noise models, and \citet{zhang2012identifiability} invertible interactions between the covariates and the noise models.  There are other methods to consider the nonlinear models such as Gaussian process regression \cite{stegle2010probabilistic}.% and kernel methods \cite{mitrovic2018causal}. 

Information theory gives an alternative view on asymmetry using Kolmogorov complexity, following the postulate that the mechanism in the causal direction should be less complex than the one in the anti-causal direction. Several papers have proposed to approximate or use certain proxies for intractable Kolmogorov complexity \cite{janzing2010causal, lemeire2006causal, daniusis2012inferring, mitrovic2018causal}.

% the use of linear and non-gaussian  models \cite{shimizu2006linear},  nonlinear additive noise models \cite{hoyer2009nonlinear}, invertible interactions between the covariates and the noise models \cite{zhang2012identifiability}, Gaussian process regression model \cite{stegle2010probabilistic} and Conditional Mean Embedding models \cite{mitrovic2018causal}.

The method that comes closest to ours is CGNN \cite{goudet2017causal}. However, our meta-CGNN method differs from CGNN in a multitude of aspects.  
\begin{enumerate}
    \item Our method employs meta learning, while CGNN only considers one dataset at a time. Hence CGNN is not able leverage similarity between datasets. A na\"ive way of training the CGNN jointly over datasets as in (Eq \ref{loss}) was analysed in our ablation study and performed poorly.
    \item  CGNN averages over $32$-$64$ separately trained generative networks per direction, which is computationally very expensive, i.e. training up to 128 models per dataset. In contrast, we merely train the model with four random initializations and average over the resulting MMDs for each dataset, thereby achieving similar results to CGNN for larger datasets and significantly better for smaller ones. 
    \item CGNN needs to be trained separately for every new dataset, which in practice can be slow as well as computationally expensive. \textit{meta-CGNN} does not need to be trained for a new dataset and can give a causal direction through a simple forward pass at the test time. This is a crucial difference which allows much faster inference once a new dataset is presented to the model. 
\end{enumerate}

Lastly, there have recently been other applications of meta learning to causal inference. MetaCI \cite{sharma2019metaci} uses a MAML-style\cite{finn2017model} (optimization based) learner but mainly deals with counterfactual causality rather than causal directionality. 

\citet{bengio2019meta} also consider a meta learning method using FCM, based on the principle that the model assuming true causal direction can be ``adapted faster'' than the one assuming the anti-causal. However, their method is designed for different settings. Firstly, the test distribution comes from a perturbation of the training distribution, which is modeled by a known parametric family. The choice of the model is not trivial for continuous domains in real-world data. Secondly, for training of neural networks, they assume to have access to a large training dataset around $3000$ for a single mechanism, which is different from our setting of small data. On another note \citet{ke2019learning} also considers meta learning in the causal setting however, they do not focus on the bivariate settings, which is our main objective.

\begin{figure*}[htp!]
    \centering
    \includegraphics[width=0.33\textwidth]{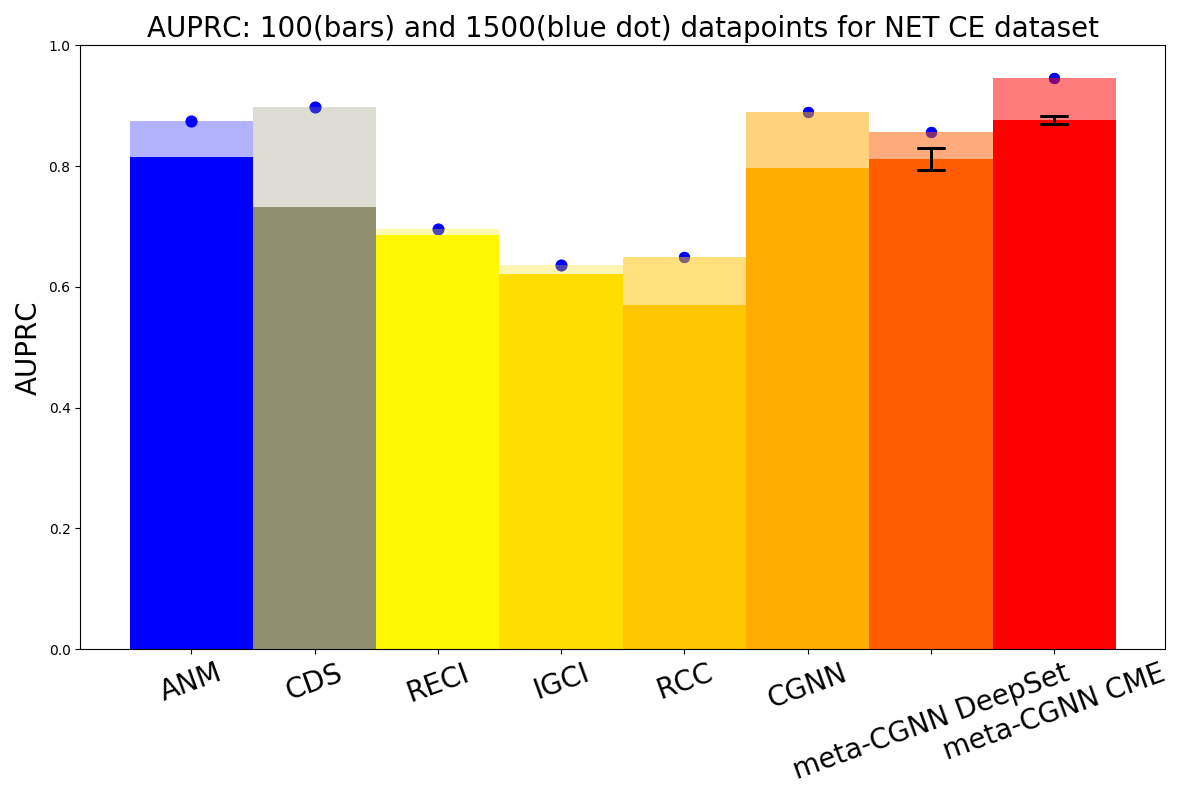}
     \includegraphics[width=0.33\textwidth]{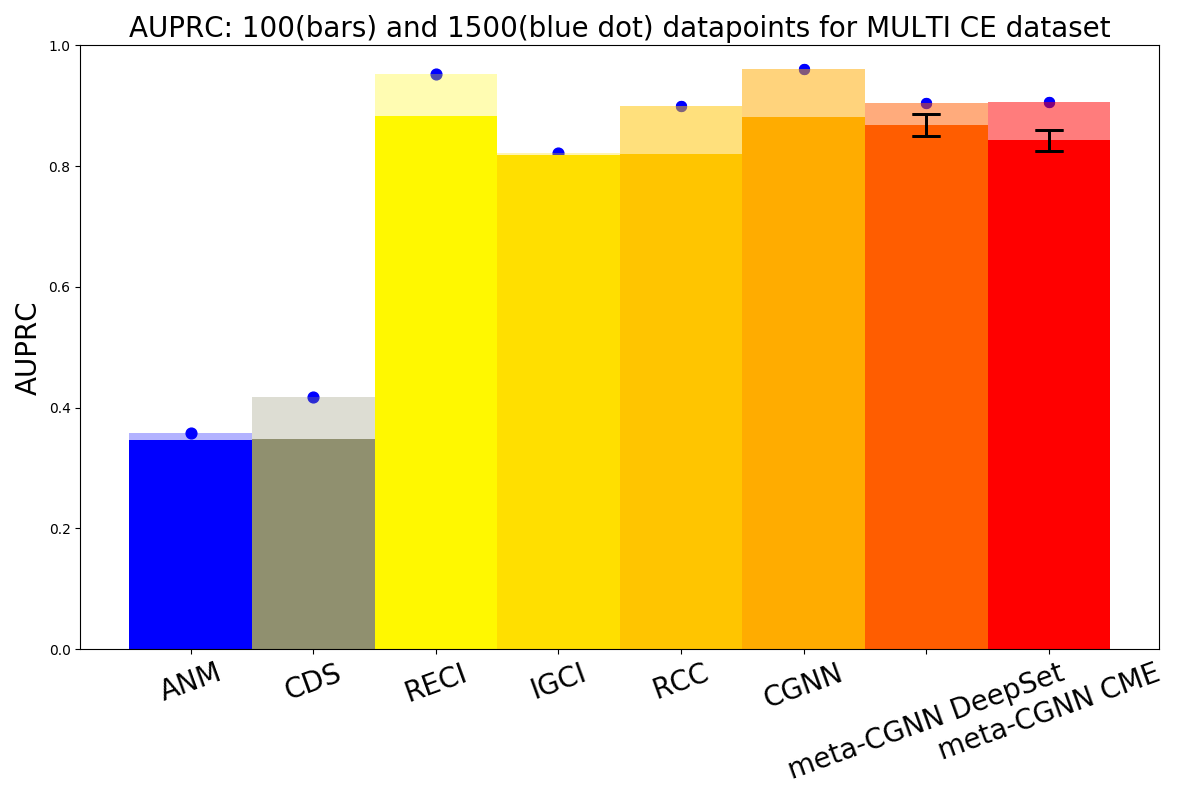} \\
     \includegraphics[width=0.33\textwidth]{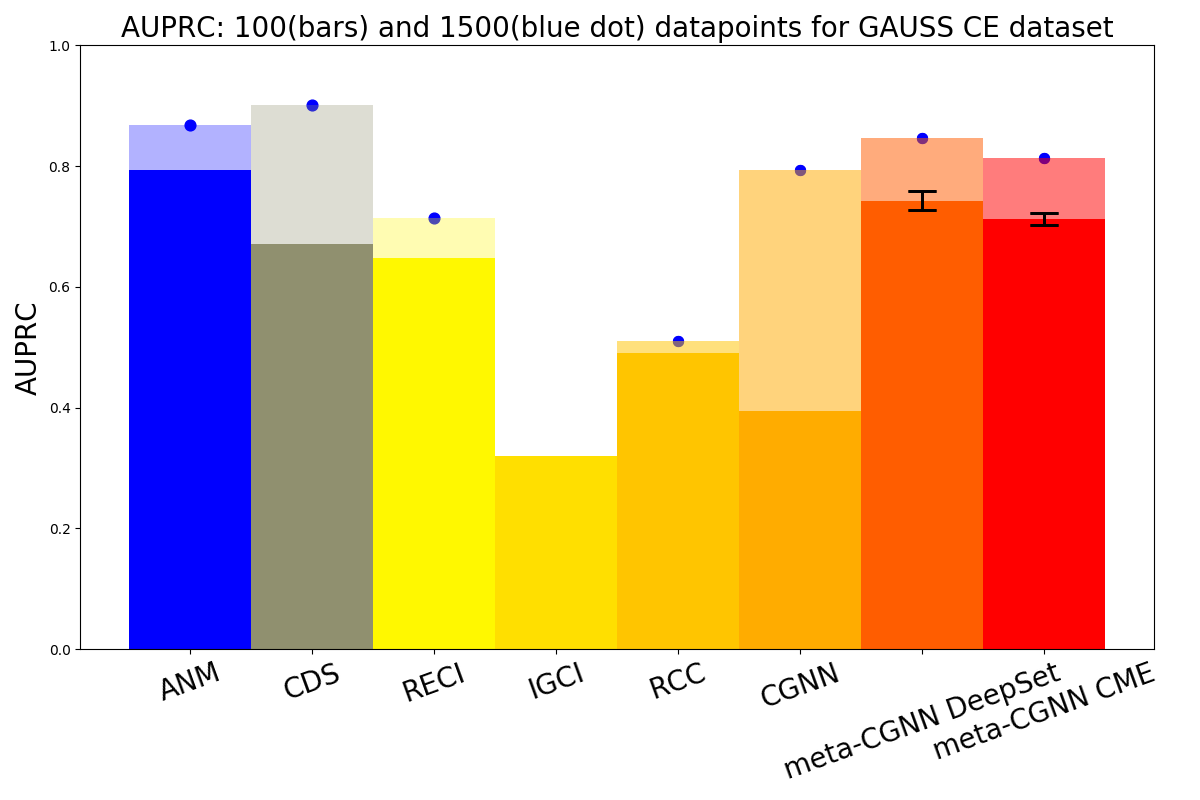}
     \includegraphics[width=0.33\textwidth]{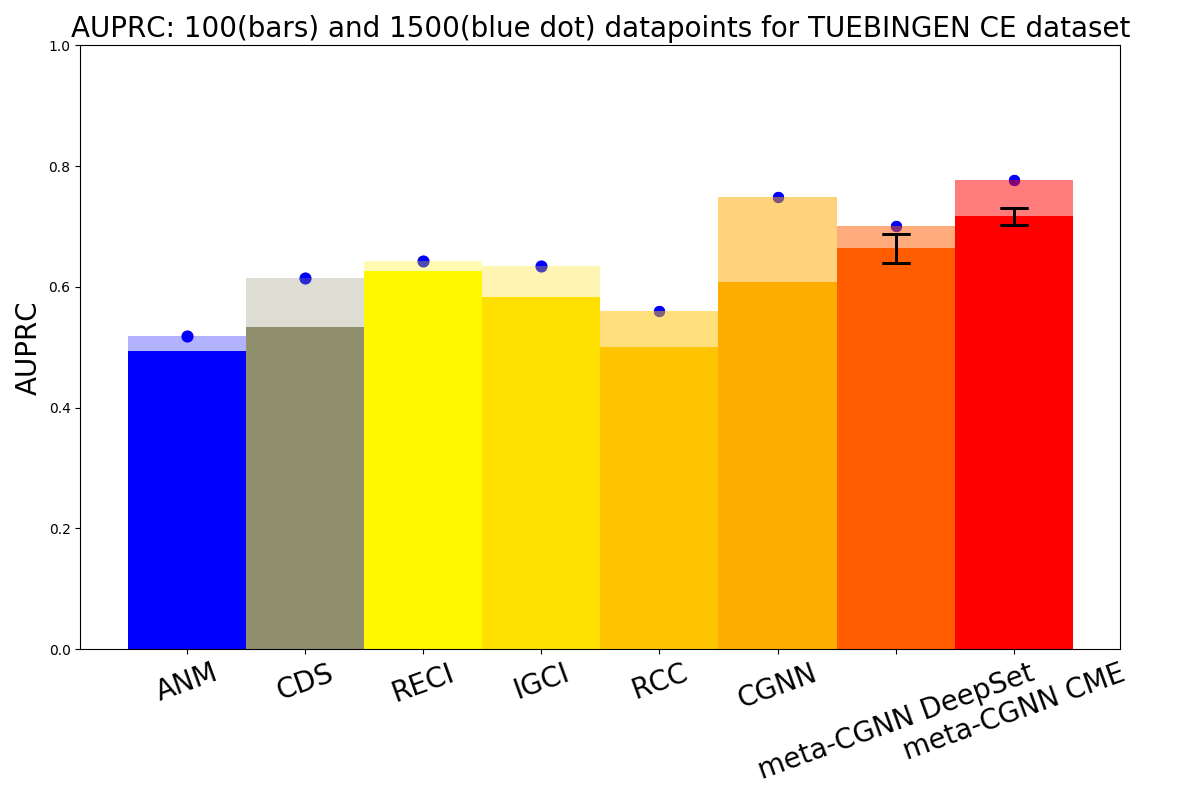}
    \caption{AUPRC for Net, Multi, Gauss and Tuebingen dataset. The thin colored bars with blue dots represent the AUPRC with 1500 datapoints, whereas the thicker barplots are with 100 datapoints. The proposed \textit{meta-CGNN} shows among the best results in all the cases. Note that the other methods show significant degradation for some datasets or small data size.}
    \label{fig:res}
\end{figure*}
\begin{table}[t]
\begin{tabular}{l|l}
\hline
Method               & Inference Time     \\ \hline
ANM                  & \textless 1sec CPU \\
CDS                  & \textless 1sec CPU \\
RECI                 & \textless 1sec CPU \\
ICGI                 & \textless 1sec CPU \\
RCC                  & \textless 1sec CPU \\
CGNN                 & 24 mins GPU        \\
meta-CGNN (DeepSets) & \textless 1min GPU \\
meta-CGNN (CME)      & \textless 1min GPU \\ \hline
\end{tabular}
\caption{Time needed during inference i.e. to determine the causal direction for a new dataset at test time. \textit{meta-CGNN} is a lot faster at inference time than CGNN, which is crucial, given that CGNN needs to be trained from scratch for every new dataset.}
\label{exp:time}
\end{table}

\begin{algorithm}[]
\label{metacme}
\caption{\textit{meta-CGNN} with CME}
\textbf{Input:} $\mathcal{D} = \{\mathcal{D}_i\}_{i=1}^N$ Datasets w/ causal
direction label\;
$\mathcal{D}^{test} = \{\mathcal{D}^{test}\}$ Datasets w/o causal direction label\;
\textbf{Output:} Causal direction for $\mathcal{D}^{test}$
\begin{algorithmic}[1]
    \FOR{$i \gets 1$ to $N$}
    \STATE{Compute CMEO: $\mathcal{C}_{Y|X}$ from $\mathcal{D}_i$ (Eq.\ref{CMEO})}
    \FOR{$j \gets 1$ to $m$}
    \STATE{Compute dataset-feature i.e $C_{i, j} = \mu_{Y|X=x_j^i}$}
    \STATE{$W^i_j := \mu(C_{i, j}) + \sigma(C_{i, j})Z^i_j$, \text{ s.t.} $Z^i_j\sim N(0,1)$}
    \STATE{$\hat{y}^i_j = D_{FL}\left((x^i_j, W^i_j); C\right)$} \text{ $\rightarrow \widehat{\mathcal{D}_i}$ with FiLM dec.}
    \ENDFOR
\ENDFOR
\STATE{$LOSS = \sum_{i=1}^N \widehat{\text{MMD}}^2(\widehat{\mathcal{D}_i}, \mathcal{D}_i)$}

\STATE{Back-propagate through \textit{LOSS} until convergence}\\
\STATE{Compute $\mathcal{M}_{xy}:=\widehat{\text{MMD}}(\mathcal{D}^{xy},\hat{\mathcal{D}}^{x\hat{y}})$ and $\mathcal{M}_{yx}:=\widehat{\text{MMD}}(\mathcal{D}^{yx},\hat{\mathcal{D}}^{y\hat{x}})$}
\STATE{Check $\mathcal{M}_{xy}<\mathcal{M}_{yx}$ or $\mathcal{M}_{xy}>\mathcal{M}_{yx}$ }
\end{algorithmic}
\end{algorithm}

\section{Experiments} \label{section:experiments}
\subsection{Synthetic Datasets}
For the synthetic experiments we use three different types of datasets taken from \citet{goudet2017causal}, each of which exhibits distinct cause-effect (CE) mechanism. The \textit{CE-Net} contains 300 cause-effect pairs with random distributions for the causes and random Neural Networks as mechanisms to create the effects. The \textit{CE-Gauss} also contains 300 data pairs, where the cause is sampled from a random mixture of Gaussians and the mechanism is drawn from a Gaussian process \cite{mooij2016distinguishing}. Lastly, \textit{CE-Multi} datasets take the cause from Gaussian distributions and the mechanisms are built using random linear and polynomial functions. They also include multiplicative and additive noise before or after the causal mechanism, making the task harder. In addition, to confirm the advantage of meta learning, we use two different data size regimes: 1500 and 100 datapoints. 

%These three types of datasets represent a wide array of CE pairs and will serve as benchmarks for our proposed method. 
%In our experiments we are mainly interested in firstly, how well and confident our method is able to distinguish the causal direction, hence similar to \cite{goudet2017causal} we use the Area Under the Precision Curve (AUPRC), we also added the accuracy in the Appendix. And secondly, how well the algorithms do with limited data. In \cite{goudet2017causal} each dataset contains $1500$ datapoints. In our experiments we will test with varying number of datapoints, from 1500 to 100. 

We measure the performance of distinguishing the causal direction using the Area Under the Precision Recall Curve (AUPRC), which is the same metric used in \cite{goudet2017causal} (accuracy is presented in the Supplementary material). With AUPRC, we are able to take into account the confidence of an algorithm, thus allowing models not to commit to a prediction if not certain. %i.e., not answering is better that giving a wrong causal direction. 
%\cite{goudet2017causal} each dataset contains $1500$ datapoints. In our experiments we will test with varying number of datapoints, from 1500 to 100. 

For \textit{meta-CGNN}, we use 100 datasets for training and the remaining 200 for testing.  We average the MMD of each dataset over 4 independent runs in order to get our final prediction. At testing time, we only need to do a simple forward pass through our model, as our model has been trained to adapt to new dataset quickly and efficiently. Hence it takes \textbf{$<$1 minute} for each new dataset at inference time. This is contrary to CGNN that needs to be trained on each new dataset separately.  Note that CGNN \cite{goudet2017causal} significantly benefits from averaging their model over multiple runs i.e. around 32-64 different runs, which takes about takes \textbf{24 minutes} per dataset. This would be infeasible without high-performance computing environments as we have hundreds of datasets to do inference on. Hence we have restricted ourselves to averaging CGNN over 12 runs in the comparisons, which is still computationally very heavy (See Table \ref{exp:time}).

Regarding architectures, we use 2 hidden layers with ReLU activation function for the FiLM, amortization and encoder network. For the decoder we use a 1 hidden layer with ReLU activation function.

As noted in \citet{goudet2017causal}, the number of hidden nodes in the decoder is very important; too small network is not able to realize the mapping properly, while too big network tends to overfit so that both directions have low MMD values. Hence for our experiments, for simplicity, we solely cross-validated over the number of decoder nodes $[5, 40]$ \cite{goudet2017causal} by leaving out a few datasets at training aside for validation. Around $40$ nodes for the 1-hidden layer was optimal for \cite{goudet2017causal}. See Supplements for further experimental details.

In addition, following \cite{goudet2017causal}, for our loss function, we use a sum over Gaussian kernel MMDs with bandwidth $\eta \in \{0.005, 0.05, 0.25, 0.5, 1, 5, 50\}$ together with an Adam optimizer \cite{kingma2014adam} and a fixed learning rate $0.01$. For the mini-batch size we fix $q$ to 10. These were parameters we fixed in the beginning and seemed to work well in our experiments.

We compare \textit{meta-CGNN} against several competing ones that have open source codes; the methods are
\begin{enumerate}
    \item Additive noise model (ANM) \cite{mooij2016distinguishing} with Gaussian process regression and HSIC test.
    \item Information Geometric Causal Inference (IGCI) \cite{daniusis2012inferring} with entropy estimator and Gaussian reference measure
    \item Conditional Distribution Similarity statistic (CDS) \cite{fonollosa2019conditional}, which analyses the variance of the conditional distributions
    \item Regression Error based Causal Inference (RECI) \cite{blobaum2019analysis}, which analyses the residual of each direction using a polynomial fit
    \item  Randomized Causation Coefficient \cite{lopez2015randomized}, which builds creates a synthetic classification problem and use CME as the feature.
    \item Causal Generative Neural Networks (CGNN) \cite{goudet2017causal}
\end{enumerate}

We use the implementation by \cite{kalainathan2019causal} which provides a GitHub repository toolbox for the above mentioned methods. In order to keep the comparisons fair, we use the same 200 data pairs as the one that our \textit{meta-CGNN} is tested on. Finally, in order to demonstrate the effect of each building block in our model, we have also conducted an ablation study, highlighting the importance of the task embeddings and FiLM layer. Further experimental details can be found in the Supplementary Material.
%
%Finally, we compute the accuracy as well as the AURPC of our method compared to previous methods.

\begin{table*}[htp]
\centering
\begin{tabular}{l|llllll}
\hline
\multirow{2}{*}{\#datapoints} & \multicolumn{2}{c}{Gauss}                                        & \multicolumn{2}{c}{Multi}                                        & \multicolumn{2}{c}{Net}                     \\ \cline{2-7} 
                               & \multicolumn{1}{l|}{FiLM}          & \multicolumn{1}{l|}{noFiLM} & \multicolumn{1}{l|}{FiLM}          & \multicolumn{1}{l|}{noFiLM} & \multicolumn{1}{l|}{FiLM}          & noFiLM \\ \hline
1500                           & \multicolumn{1}{l|}{\textbf{0.78}} & \multicolumn{1}{l|}{0.73}   & \multicolumn{1}{l|}{\textbf{0.75}} & \multicolumn{1}{l|}{0.46}   & \multicolumn{1}{l|}{\textbf{0.70}} & 0.55   \\
500                            & \multicolumn{1}{l|}{\textbf{0.80}} & \multicolumn{1}{l|}{0.72}   & \multicolumn{1}{l|}{\textbf{0.73}} & \multicolumn{1}{l|}{0.49}   & \multicolumn{1}{l|}{\textbf{0.70}} & 0.50   \\
100                            & \multicolumn{1}{l|}{\textbf{0.73}} & \multicolumn{1}{l|}{0.64}   & \multicolumn{1}{l|}{\textbf{0.72}} & \multicolumn{1}{l|}{0.63}   & \multicolumn{1}{l|}{\textbf{0.72}} & 0.50   
\end{tabular}
\caption{Accuracy of the proposed method with DeepSet embedding with and without the FiLM layer}
\label{exp:ablation}
\end{table*}

\subsection{Tuebingen Cause-Effect dataset}
As a real-world example, we use the popular Tuebingen benchmark \cite{mooij2016distinguishing}, from which we take 99 bivariate datasets. We use a similar setup as in the synthetic experiments, and the only difference is that we employ 5-fold cross-validation for training and testing, i.e. We train on 4 of the folds and test on the last one. We repeat this procedure such that each fold has been the test fold at one point, while the remaining were acting as training. 
That way we obtain a prediction on each of the 99 datasets. We repeat this 3 times with different random splits and report the results in Figure \ref{fig:res}, where we also check the performance on decreasing size of samples in a dataset.
%and we can see from the Figure \ref{fig:res}, that using the \textit{meta-CGNN}, we are able to maintain high performance even when decreasing the dataset size. 
% \textcolor{red}{[The following should be explained in the next Section.] This does not seem to hold true for the standard CGNN, which sees a significant drop in performance when faced with less data. We have added more experiments in the Appendix when setting the dataset size to at most 1500, 1000, 500, 100 and see the same trend.}
% In addition, interestingly CME embeddings seems to have more conservative predictions looking at the AUPRC compared to the DeepSets approach of embedding the tasks.

% \begin{figure}[htp]
%     \centering
%     \includegraphics[width=0.49\textwidth]{res_figs2/Tueb_acc_med2.png}
%     \includegraphics[width=0.49\textwidth]{res_figs2/Tueb_auprc_med2.png}
%     \caption{Accuracy/ AURPC for standard CGNN, meta-DeepSet, meta-CME}
%     \label{fig:galaxy}
% \end{figure}

\subsection{Results}
Figure 2 illustrates how both our \textit{meta-CGNN} algorithms are the only ones that can retain high AUPRC across different datasets, as well as dataset sizes, i.e. 1500 and 100 datatpoints. \textbf{NOTE:} \textit{meta-CGNN} performs well in all the small dataset settings, while remaining computationally more efficient at inference time, in contrast to CGNN, which has significantly worse result in the small data for \textit{CE-Gauss} and \textit{CE-Tueb} dataset and requires training the whole model at test time. Another important point to notice is that similarly to CGNN, \textit{meta-CGNN} does not have any assumptions on the data and hence can be used in a variety of datasets, while retaining good performance. This does not hold for other methods.

%Even though \textit{meta-CGNN} does not perform the best in every dataset, the important thing to notice is that similarly to CGNN it does not have any assumptions on the data and hence can be used in a variety of dataset. In addition, contrary to CGNN, \textit{meta-CGNN} not only works well on all the datasets, but also works

Algorithms such as ANM \cite{mooij2016distinguishing}, for example, seems to do reasonably well on the \textit{CE-Net} and \textit{CE-Gauss} dataset, but completely fails in the \textit{CE-Multi} and, more importantly, on the real-world Tuebingen dataset. This occurs mainly because of the strict assumptions which ANM imposes on the FCM. Similarly, RECI which performs well on the \textit{CE-Multi}, but not on the \textit{CE-Net}, \textit{CE-Gauss} and the Tuebingen dataset. Similar situation holds true for the remaining competing methods. 

% The only other method that also retains high performance across datasets is CGNN. However, in particular \textit{CE-Gauss}, \textit{CE-Net} and the Tuebingen dataset, we can see substantial decrease  in AUPRC, as well as in accuracy (see Supplement), when only given small data compared to \textit{meta-CGNN}. This illustrates the data dependence of CGNN and highlights the benefits of our approach.

Our proposed method is amongst the top performing ones and is consistently doing well for both 1500 and 100 datapoints settings. These results illustrate that \textit{meta-CGNN} is able to retain high performance, even when faced with small data, by leveraging the meta-learning setting. In order to understand the components of \textit{meta-CGNN} better we conducted an ablation study  in the next section.

% However, there is an outlier in our results, which is \textit{meta-CGNN} with CME task representation which performs poorly on the \textit{CE-Gauss} dataset (while still outperforming CGNN in the small data regime). We hypothesise that this is due the multi-modality in the source distribution which may be difficult for the CME to capture. 

% Another observation to make is that \textit{meta-CGNN} CME and DeepSets seem to have comparable performance (except on the Gauss dataset), with the CME version winning out on the \textit{CE-Net} and Tuebingen dataset. This is interesting, in two ways. firstly, \textit{meta-CGNN} CME does not need to train its task embeddings as they are given through the CMEO using a characteristic kernel such as the RBF kernel. Secondly, the CMEO can be pre-computed before training, which makes it computationally efficient. 

Lastly, we want to emphasise that \textit{meta-CGNN} does not need to be trained for each test dataset, but instead only needs a forward pass through the generative model to determine the causal direction, which makes it vastly more computationally efficient than CGNN, while attaining higher performance (see Table \ref{exp:time}).

\subsection{Ablation study}
\label{sec:ablation}
In this section, we study the necessity of task embedding in our model and show that not every model can be trained in this joint dataset fashion. To this end, we used a CGNN \cite{goudet2017causal} that has been trained exactly like our \textit{meta-CGNN}, i.e. \textit{meta-CGNN} where we removed the amortization network and the FiLM layer. In this case, only the decoder was trained jointly across the datasets and we see that this na\"ive version of extending CGNN does not perform well, with accuracy hovering around $55$-$60\%$, which are close to chance level, for the Tuebingen dataset with at most 1500 datapoints per dataset. It behaves similarly on the synthetic data with a \textbf{chance level accuracy} for \textit{CE-Net} dataset with 1500 datapoints. 

% Note that we have also tried instead of using a FiLM layer to just concatenate $[x, z, C_i]$ into one vector for the decoder network. This resulted in equal or slightly worse performance than using the FiLM layer.

Next, we investigate the importance of the FiLM layer and run the experiments from our experiment section using DeepSets embeddings, with and without the FiLM layer while keeping the amortization network. We see from Table \ref{exp:ablation} a significant performance boost when using the FiLM as it allows the decoder at test time more flexibility to adapt, instead of relying solely on the amortization network. This shows the importance of each component in the proposed architecture and that adapting cause-effect methods into the meta learning setting is a non-trivial task. We see similar trends for \textit{meta-CGNN} using CME.
Lastly, we have also tried to ablate for the amortization network and noticed a similar/slightly higher performance with the amortization network in \textit{meta-CGNN}.

\section{Conclusion}
We introduced a novel meta learning algorithm for cause-effect inference which performs well even in the small data regime, in sharp contrast to the existing methods. By leveraging a dataset-feature extractor that can be learned during training, we are able to efficiently adapt our model at test time to new previously unseen datasets by using amortization and FiLM layers. We also demonstrate the utility of using conditional mean embeddings, as they allow us to capture the distribution and adapt the model at test time.

In addition, we extended the framework of Causal Generative Neural Network (CGNN) \cite{goudet2017causal} by learning a single generative network, readily adaptable for new datasets, vastly alleviating the computation burden of CGNNs. In particular, instead of having to train multiple models on each dataset separately, our proposed methods are able to achieve similar or better performance than existing methods with simple forward passes through our generative network at test time. 

Recently there has been an increase in interest in causality topics in reinforcement learning \cite{zhu2019causal, buesing2018woulda,dasgupta2019causal}, where the proposed methods may also be applicable. Assuming that we have a skeleton graph of the relevant quantities in the RL model, \textit{meta-CGNN} could be used to efficiently infer causal direction of unoriented edges. %Hence allowing for more efficient training. 

\section{Acknowledgements}
We would like to thank Pengzhou (Abel) Wu for interesting discussions. JFT is supported by the EPSRC and MRC through the OxWaSP CDT programme EP/L016710/1. KF is supported by SPS KAKENHI 18K19793 and JST CREST JPMJCR2015.

\bibliography{ref}
\newpage
\appendix

\section{Appendix}
\subsection{Meta Learning}
Meta learning is an ever growing area in machine learning as it allows a model to extract information from similar problems/datasets and use this prior information on new unseen datasets. This is achieved, by sharing statistical strength across several causal inference problems. Standard methods usually require a lot of prior information to be useful in small data setting, i.e. knowing it is a linear model etc. Meta learning however learns this prior information through the meta learning training phase, where during training the model is presented with several small datasets, that allows the model to perform well on new unseen dataset.

Meta learning and multi-task learning differ in the following way:
\begin{itemize}
    \item "\textit{The meta-learning problem: Given data/experience on previous tasks, learn a \textbf{new task} more quickly and/or more proficiently}" (Finn, 2019)
    \item "\textit{The multi-task learning problem: Learn all of the tasks more quickly or more proficiently than learning them independently}"(Finn, 2019)
\end{itemize}
In our setting, we are interested in learning the causal direction on new unseen datasets, given a set of datasets where we know the causal direction. This allows us to make efficient use of the information across all the datasets, which is contrary to most cause-effect methods that treat datasets independently and have to be retrained for each new dataset.

\subsection{Ablation study}
\label{sec:ablation}
In this section, we study the necessity of task embedding in our model and show that not every model can be trained in this joint dataset fashion. To this end, we used a CGNN \cite{goudet2017causal} that has been trained exactly like our \textit{meta-CGNN}, i.e. \textit{meta-CGNN} where we removed the amortization network and the FiLM layer. In this case, only the decoder was trained jointly across the datasets and we see that this na\"ive version of extending CGNN does not perform well, with accuracy hovering around $55$-$60\%$, which are close to chance level, for the Tuebingen dataset with at most 1500 datapoints per dataset. It behaves similarly on the synthetic data with a \textbf{chance level accuracy} for \textit{CE-Net} dataset with 1500 datapoints. 

% Note that we have also tried instead of using a FiLM layer to just concatenate $[x, z, C_i]$ into one vector for the decoder network. This resulted in equal or slightly worse performance than using the FiLM layer.

Next, we investigate the importance of the FiLM layer and run the experiments from our experiment section using DeepSets embeddings, with and without the FiLM layer while keeping the amortization network. We see from Table \ref{exp:ablation} a significant performance boost when using the FiLM as it allows the decoder at test time more flexibility to adapt, instead of relying solely on the amortization network. This shows the importance of each component in the proposed architecture and that adapting cause-effect methods into the meta learning setting is a non-trivial task. We see similar trends for \textit{meta-CGNN} using CME.
Lastly, we have also tried to ablate for the amortization network and noticed a similar/slightly higher performance with the amortization network in \textit{meta-CGNN}.

\section{Additional details on experiments setup}
In this section of the Appendix we will explain the experimental setup in more detail. For all our synthetic datasets, we have 300 datasets. We then use 100 pairs to train our meta-CGNN algorithm and test on the remaining 200. For the standard methods, we only consider the 200 testing datasets as they cannot incorporate information from different datasets. Each dataset then has either 1500 or 100 datapoints. 

Note that for the Tuebingen dataset contains 99 bivariate datasets. We use the same setup as in \cite{goudet2018learning}, which means that datasets have at most 1500 or at most 100 datapoints. We use a similar setup as in the synthetic experiments, and the only difference is that we employ 5-fold cross-validation for training and testing. To be concrete, we train on 4 of the folds and test on the last one. We repeat this procedure such that each fold has been the test fold at one point, while the remaining were acting as training. That way we can obtain fair estimates of the directions. We repeat these experiments 3 times with different splits.

Next we provide the neural network architecture for meta-CGNN as well as hyper-parameters. For simplicity, just as in \cite{goudet2018learning}, we use a fixed learning rate of $0.01$, Adam optimizer, and up to 500 epochs. We used a mini-batch size of 10, i.e. 10 datasets for each gradient update. In terms of the neural network architectures we used a 2-hidden layer network for the encoder, FiLM and amortization network. For the decoder we follow \cite{goudet2018learning} and use a 1-hidden layer NN. All activations used are ReLU.

For the encoder network (meta-CGNN DeepSet), we fix $[40, 10]$ hidden nodes per layer. For the hyper-parameters $\lambda$ of the Conditional Mean Embedding Operator (CMEO) as well as lengthscale $l$ for the Random Fourier Features embedding we used $\lambda=1$ and $l=1$. These were the default settings and worked well. In the future we will cross-validate over these to see if there are any performance gains.

For the amortization network we have $[40, 20]$ hidden nodes layer and output $\mu_z$ and $\sigma_z$ which are both $1D$. In our case the latent variable $Z$ is always one dimensional. For the FiLM layer we have $[40, 40]$ hidden nodes and output $2$x decoder hidden nodes. This is because we need a shift and a scale parameter for the hidden layer of the decoder. In terms of the decoder, we cross-validate and choose either $40$ or $5$ hidden nodes.
The reason we chose values around $40$ was because we took \cite{goudet2018learning} as a starting point and it worked well, around $40$.

\begin{table*}[htp]
\centering
\begin{tabular}{l|llllll}
\hline
\multirow{2}{*}{\#datapoints} & \multicolumn{2}{c}{Gauss}                                        & \multicolumn{2}{c}{Multi}                                        & \multicolumn{2}{c}{Net}                     \\ \cline{2-7} 
                               & \multicolumn{1}{l|}{FiLM}          & \multicolumn{1}{l|}{noFiLM} & \multicolumn{1}{l|}{FiLM}          & \multicolumn{1}{l|}{noFiLM} & \multicolumn{1}{l|}{FiLM}          & noFiLM \\ \hline
1500                           & \multicolumn{1}{l|}{\textbf{0.78}} & \multicolumn{1}{l|}{0.73}   & \multicolumn{1}{l|}{\textbf{0.75}} & \multicolumn{1}{l|}{0.46}   & \multicolumn{1}{l|}{\textbf{0.70}} & 0.55   \\
500                            & \multicolumn{1}{l|}{\textbf{0.80}} & \multicolumn{1}{l|}{0.72}   & \multicolumn{1}{l|}{\textbf{0.73}} & \multicolumn{1}{l|}{0.49}   & \multicolumn{1}{l|}{\textbf{0.70}} & 0.50   \\
100                            & \multicolumn{1}{l|}{\textbf{0.73}} & \multicolumn{1}{l|}{0.64}   & \multicolumn{1}{l|}{\textbf{0.72}} & \multicolumn{1}{l|}{0.63}   & \multicolumn{1}{l|}{\textbf{0.72}} & 0.50   
\end{tabular}
\caption{Accuracy of the proposed method with DeepSet embedding with and without the FiLM layer}
\label{exp:ablation}
\end{table*}

\section{AUPRC and Accuracy of the methods}

Here we present the AURPC (area under the precision recall curve) as well as accuracy of the proposed methods as well as competing methods. Note that we report the weighted accuracy for the Tuebingen dataset as in \cite{mooij2016distinguishing} to account for similarities in the datasets. We note that the accuracy follows a similar trend to the AUPRC, however we opted to show the AUPRC in the main text as it is more indicative of the confidence for each prediction. AUPRC also allows us to take into account the fact that a method might be uncertain about its prediction, i.e. it is better to give no answer than a wrong answer.
\begin{figure*}[htp]
    \centering
    \includegraphics[width=0.35\textwidth]{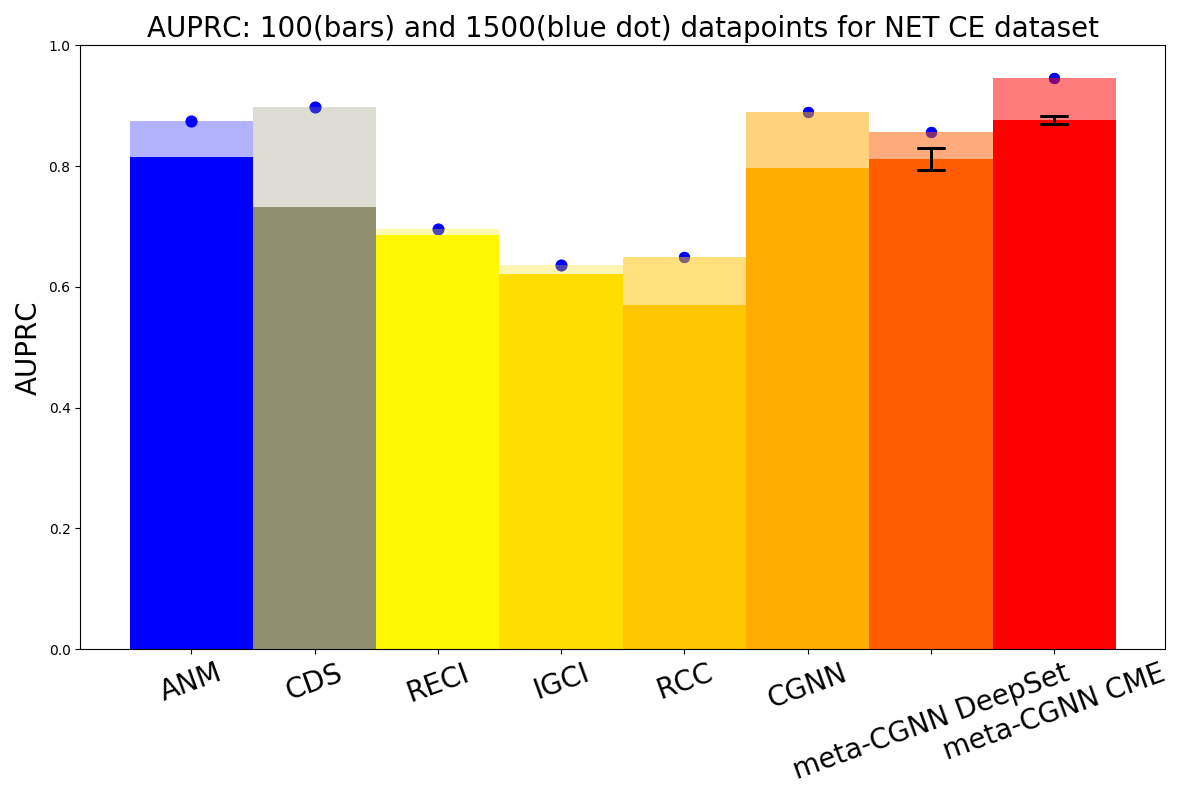}
     \includegraphics[width=0.35\textwidth]{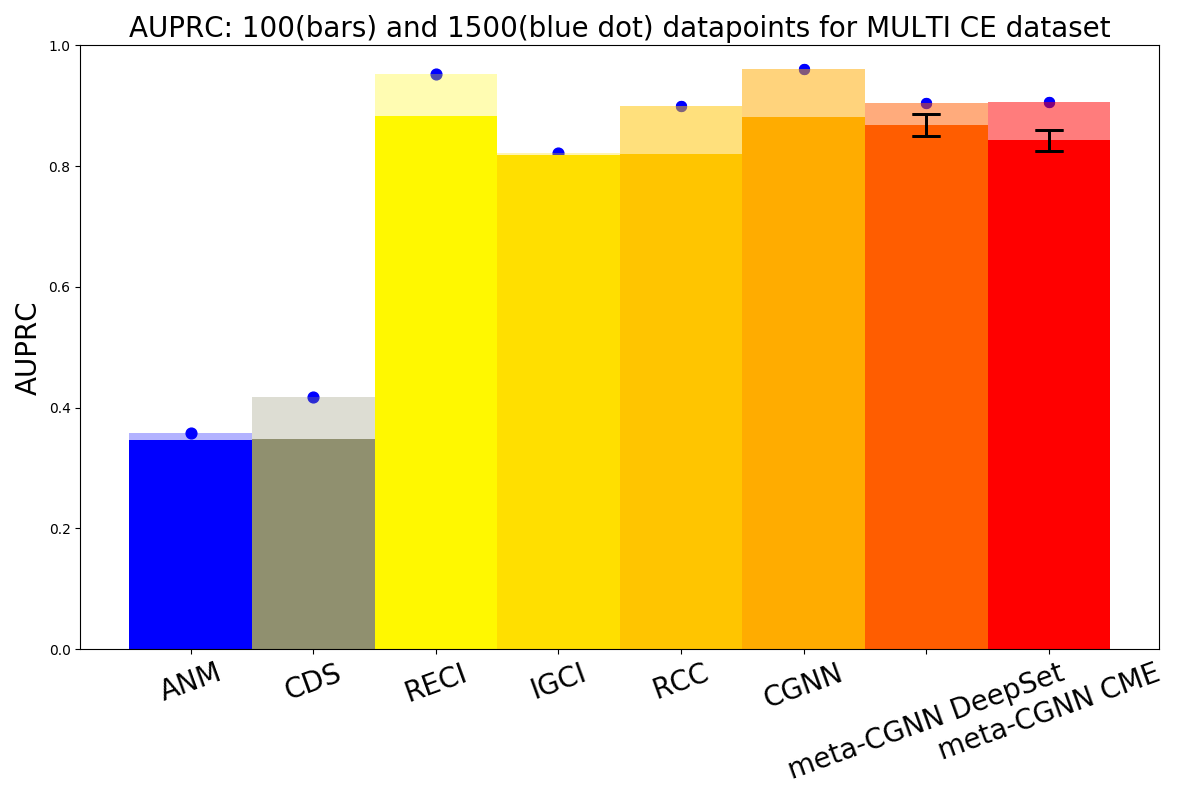} \\
     \includegraphics[width=0.35\textwidth]{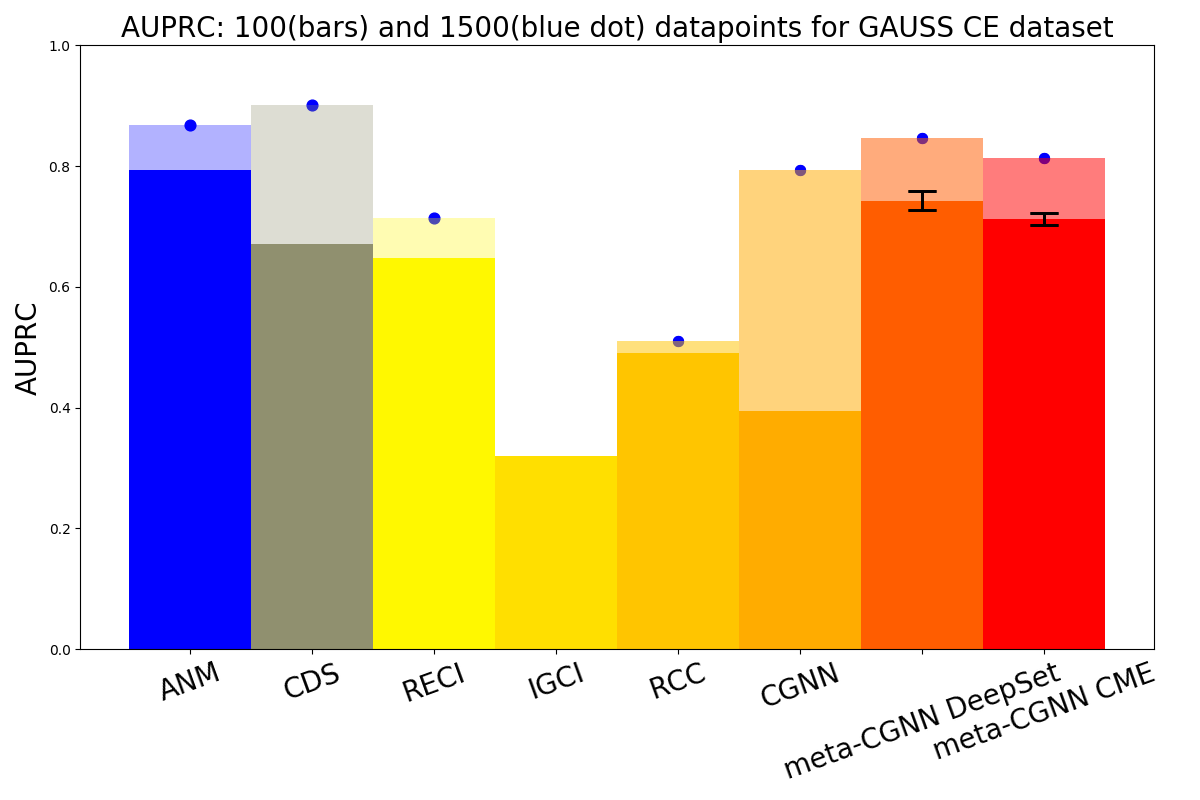}
     \includegraphics[width=0.35\textwidth]{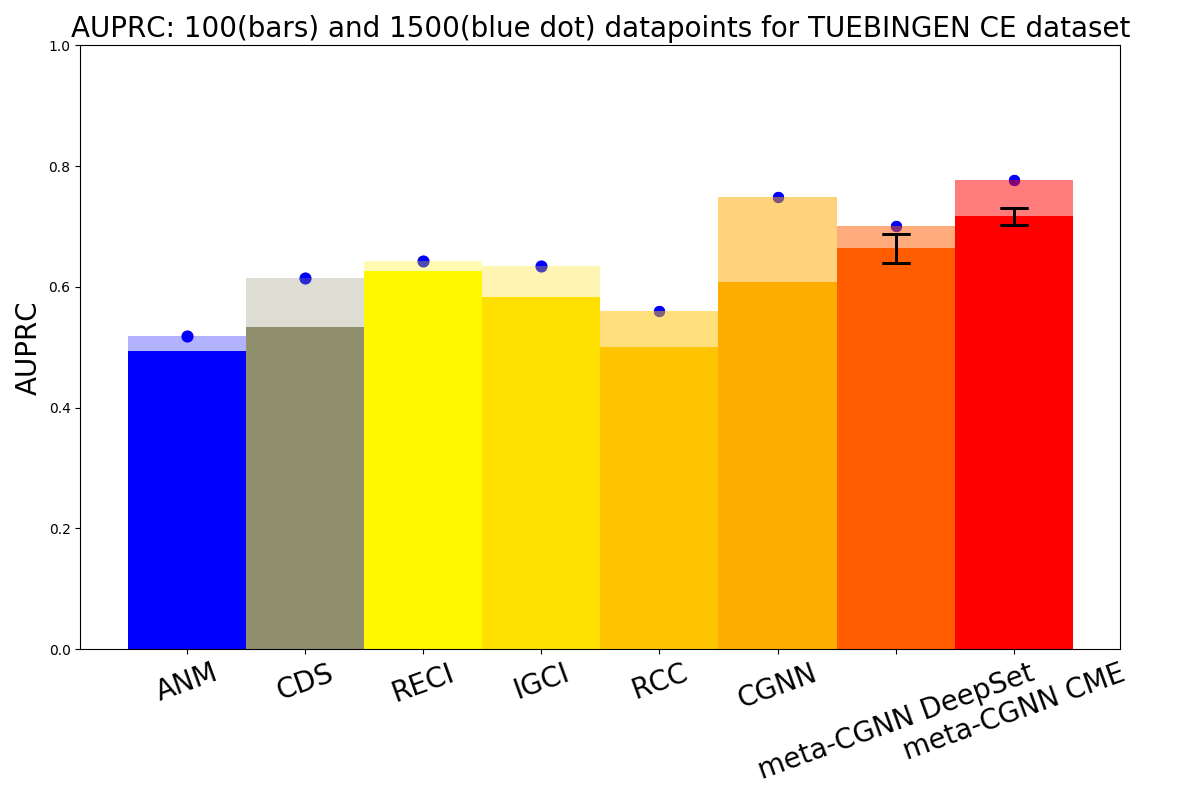}
    \caption{AUPRC for Net, Multi, Gauss and Tuebingen dataset. The blue dots represent the AUPRC with 1500 datapoints, whereas the barplots are with 100 datapoints. Note that the proposed meta-CGNN is in most cases significantly better that standard CGNN, which is trained individually on each dataset. In addition, meta-CGNN even though not the best on each dataset remains comparable across different datasets and sees the smallest drop in performance when reducing the dataset size.}
    \label{fig:res}
\end{figure*}

\begin{figure*}[htp]
    \centering
    \includegraphics[width=0.35\textwidth]{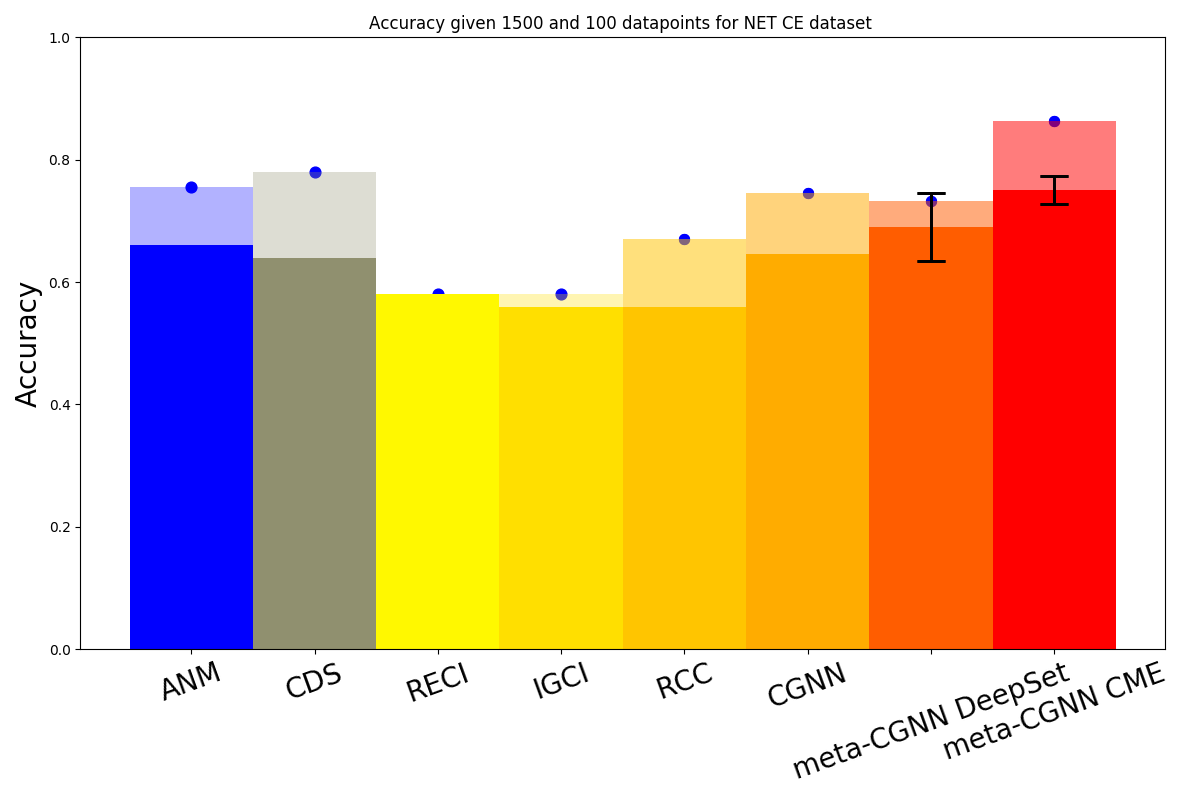}
     \includegraphics[width=0.35\textwidth]{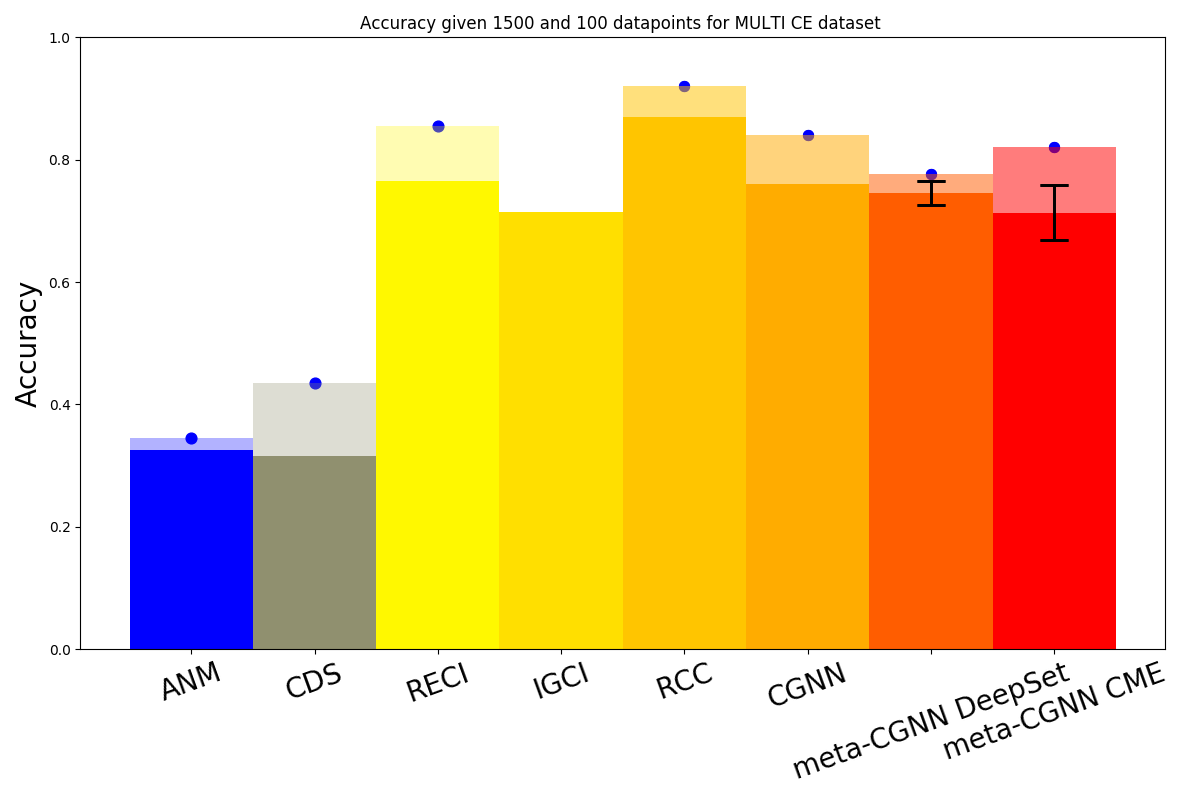} \\
     \includegraphics[width=0.35\textwidth]{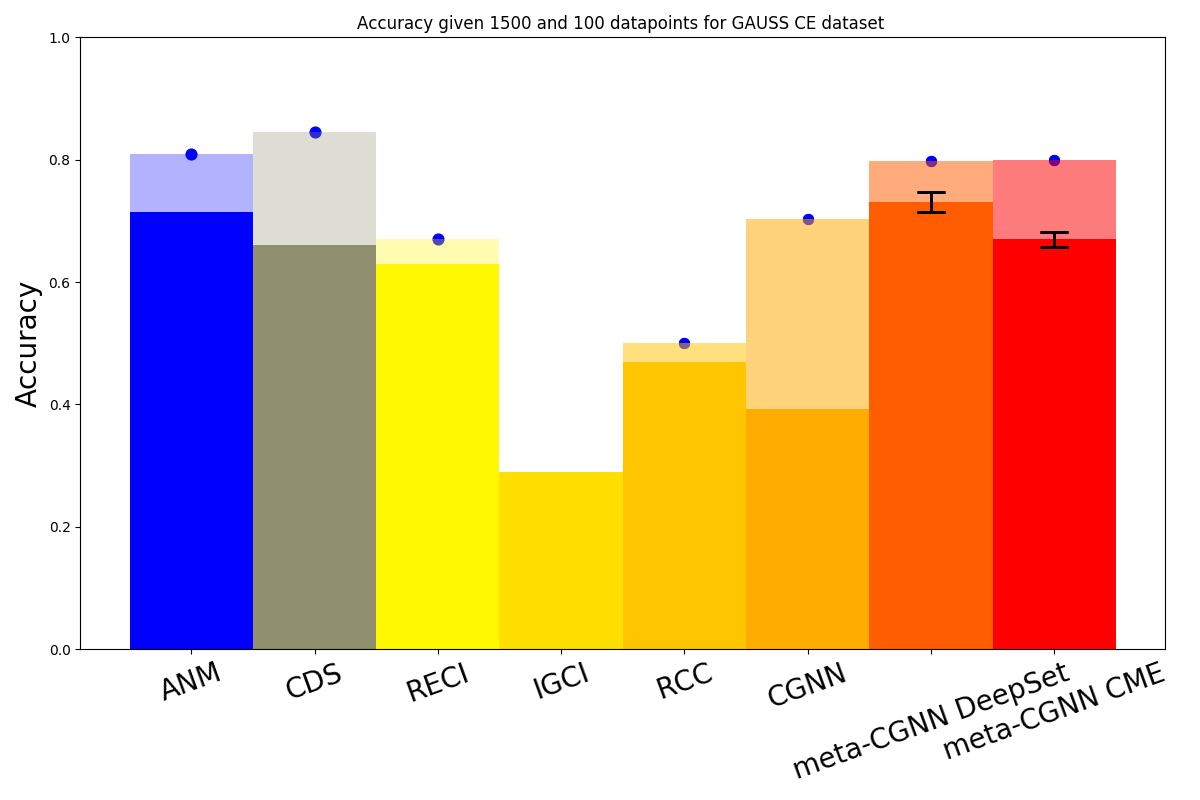}
     \includegraphics[width=0.35\textwidth]{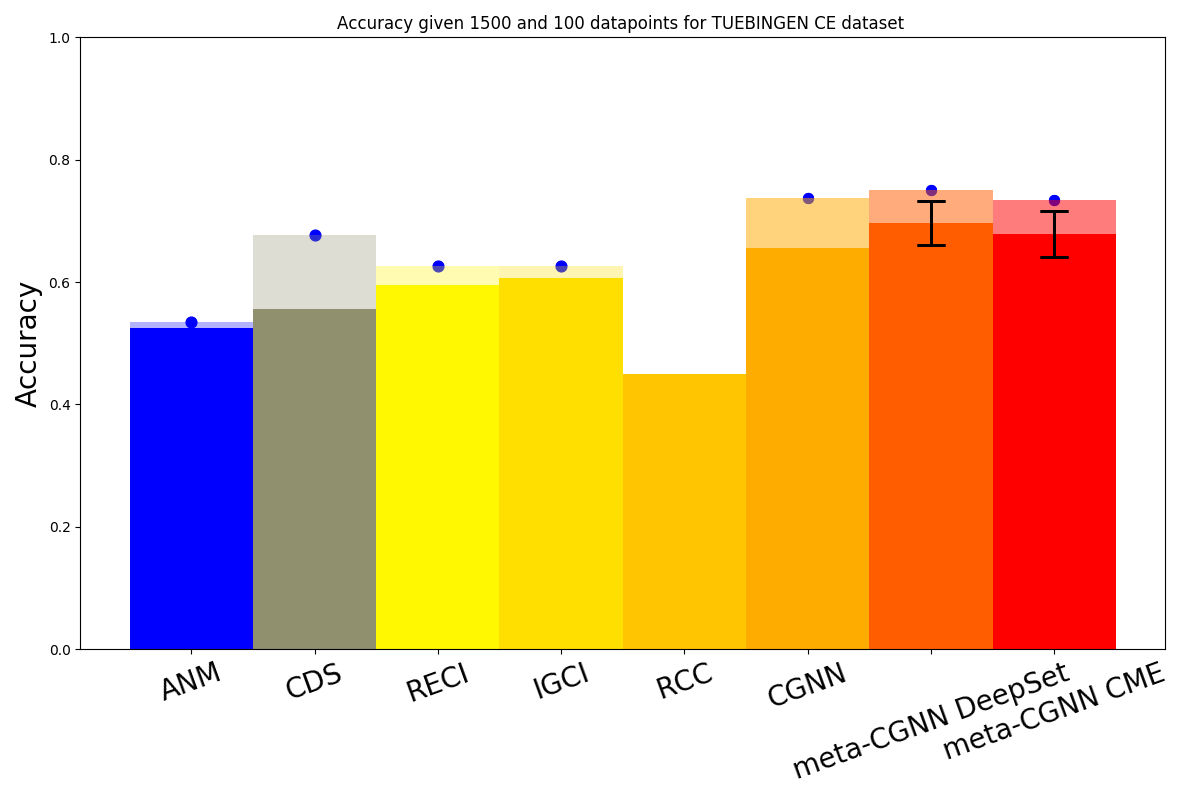}
    \caption{ACC for Net, Multi, Gauss and Tuebingen dataset. The blue dots represent the ACC with 1500 datapoints, whereas the barplots are with 100 datapoints. Note that the proposed meta-CGNN is in most cases significantly better that standard CGNN, which is trained individually on each dataset. In addition, meta-CGNN even though not the best on each dataset remains comparable across different datasets and sees the smallest drop in performance when reducing the dataset size.}
    \label{fig:res}
\end{figure*}

\section{Note on competing methods and RCC}
For the competing methods, we used the excellent \emph{Causal discovery toolbox} by \cite{kalainathan2019causal}, which is provided in an open source Github repository, with all the popular methods implemented. We have done minimal changes to the algorithms as one can easily streamline the experiments by just inserting the datasets into their framework. The only thing we changed was to use $12$ instead of $32$ or $64$ different models for CGNN \cite{goudet2018learning} and kept the hidden layer size as recommended in the paper around $40$ \cite{goudet2018learning_openreview}. The rest remained the same as described in the CGNN paper. The results we got on the $1500$ datapoints are comparable to the one noted in the paper with a slight degradation of the performance due to less model averaging, i.e. $12$ instead $32$ or $64$, which was not feasible due to our computational constraints.

In addition, we want to note that CGNN requires about 24 min per dataset at testing time, whereas our method can do inference in less than $1$min. This is a big computational saving at test time while retaining similar or better performance.

Lastly, we also investigated the performance in particular the performance of RCC (randomized causation coefficient) method \cite{lopez2015randomized} in our setting. RCC also uses the formalism of kernel mean embeddings (approximated using random features) in order to build a classifier for causal direction. However, it is data hungry and typically requires a lot of generated synthetic data to train the classifier, as noted in \cite{wu2020causal}. %In addition, creating the training data is also a non-trivial task. 
Using again the toolbox by \cite{kalainathan2019causal}, we use the same procedure as in meta-CGNN, where 100 of the the synthetic datasets are used for training and remaining ones for testing. For the Tuebingen dataset, we again perform a 5-fold cross-validation, repeating the experiment three times, and reporting mean and standard deviation in Table \ref{tab:rcc}. The performance of RCC is poor except for the \textit{CE-Multi} dataset where most other methods also did reasonably well, thus showing is that using CME alone does not solve the problem.

% In particular, RCC does especially bad on the \textit{CE-Gauss} dataset which suggests that it might be challenging to capture the structure of this dataset using kernel mean embeddings, pointing to the possible explanation of why meta-CGNN CME is doing worse on that dataset.
\begin{table*}[]
\centering
\begin{tabular}{l|l|l|l|l}
ACC  & Gauss & Multi & Net  & Tueb            \\ \hline
1500 & 0.50  & 0.92  & 0.67 & 0.39 $\pm$ 0.02 \\
100  & 0.47  & 0.87  & 0.56 & 0.45 $\pm$ 0.03
\end{tabular}
\end{table*}

\begin{table*}[]
\centering
\begin{tabular}{l|l|l|l|l}
AUPRC & Gauss & Multi & Net  & Tueb             \\ \hline
1500  & 0.51  & 0.90  & 0.65 & 0.56 $\pm$ 0.01 \\
100   & 0.49  & 0.82  & 0.57 & 0.50 $\pm$ 0.03 
\end{tabular}
\caption{Results for RCC method. As we can see RCC also struggles with \textit{CE-Gauss} dataset and seems to reasonable well on the \textit{CE-Multi} dataset.}
\label{tab:rcc}.
\end{table*}

\end{document}